\documentclass[acmtog,authorversion]{acmart}

\setcitestyle{square}

\usepackage{amsmath,amsthm,amssymb}
\usepackage{graphicx}
\usepackage{textcomp}
\usepackage{wrapfig}
\usepackage{subfig}
\usepackage{color}
\usepackage{xspace}
\usepackage{overpic}
\usepackage{subfig}
\usepackage{wrapfig}
\usepackage{enumitem}
\usepackage{mathrsfs}
\usepackage{multirow}

\definecolor{turquoise}{cmyk}{0.65,0,0.1,0.1}
\definecolor{purple}{rgb}{0.65,0,0.65}
\definecolor{dark_green}{rgb}{0, 0.4, 0}
\definecolor{dark_blue}{rgb}{0, 0, 0.4}
\definecolor{orange}{rgb}{0.6, 0.3, 0.0}
\definecolor{red}{rgb}{0.8, 0.2, 0.2}
\definecolor{brown}{rgb}{0.5, 0.16, 0.16}
\newcommand{\kx}[1]{{\color{dark_green}#1}}

\newcommand{\rz}[1]{{\color{black}#1}}

\newcommand{\revise}[1]{{\color{black}#1}}

\renewcommand{\textrightarrow}{$\rightarrow$}

\setcopyright{acmcopyright}

\begin{document}
\title{LOGAN: Unpaired Shape Transform in Latent Overcomplete Space}

\author{Kangxue Yin}
\affiliation{%
	\institution{Simon Fraser University}
}
\author{Zhiqin Chen}
\affiliation{%
	\institution{Simon Fraser University}
}
\author{Hui Huang}
\affiliation{%
	\institution{Shenzhen University}
}
\author{Daniel Cohen-Or}
\affiliation{%
	\institution{Tel Aviv University}
}
\author{Hao Zhang}
\affiliation{%
	\institution{Simon Fraser University}
}
\renewcommand\shortauthors{K. Yin, Z. Chen, H. Huang, D. Cohen-Or,  H. Zhang}

\acmJournal{TOG}
\acmYear{2019}\acmVolume{38}\acmNumber{6}\acmArticle{198}\acmMonth{11} \acmDOI{10.1145/3355089.3356494}

\begin{abstract}
We introduce LOGAN, a deep neural network aimed at learning {\em general-purpose\/} shape transforms 
from {\em unpaired\/} domains. The network is trained on two sets of shapes, e.g., tables and chairs, while 
there is neither a pairing between shapes from the domains as supervision nor any point-wise correspondence 
between any shapes. Once trained, LOGAN takes a shape from one domain and transforms it into the 
other. Our network consists of an autoencoder to encode shapes from the two input domains into a {\em 
common latent space\/}, where the latent codes {\em concatenate multi-scale\/} shape features, resulting in
an {\em overcomplete\/} representation. The translator is based on a generative adversarial network (GAN), operating
in the latent space, where an adversarial loss enforces cross-domain translation while a {\em feature
preservation loss\/} ensures that the right shape features are preserved for a natural shape transform.
We conduct ablation studies to validate each of our key network designs and demonstrate 
superior capabilities in unpaired shape transforms on a variety of examples over baselines and
state-of-the-art approaches. We show that LOGAN is able to learn what shape features to 
preserve during shape translation, either local or non-local, whether content or style,
depending solely on the input domains for training.

\end{abstract}

%
%





\begin{teaserfigure}
  \centering
  \includegraphics[width=\linewidth]{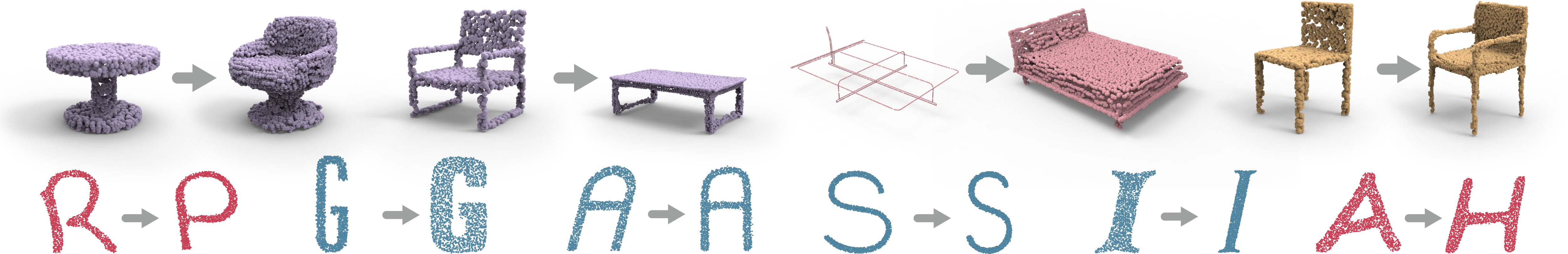}
	\caption{We present LOGAN, a deep neural network which learns {\em general-purpose\/} shape transforms from {\em unpaired\/} domains. By altering only the two input data domains for training, without changing the network architecture or any hyper-parameters, LOGAN can transform between chairs and tables, from cross-sectional profiles to surfaces, as well as adding arms to chairs. It can also learn both style-preserving content transfer (letters $R\rightarrow P$,  $A\rightarrow H$, in different font styles) and content-preserving style transfer (wide to narrow $S$, thick to thin $I$, thin to thick $G$, and italic to non-italic $A$.)}
	\label{fig:teaser}
\end{teaserfigure}

\maketitle




\section{Introduction}
\label{sec:intro}

Shape transform is one of the most fundamental and frequently encountered problems in computer graphics
and geometric modeling. With much interest in geometric deep learning in the graphics community today,
it is natural to explore whether a machine can learn shape transforms, particularly under the {\em unsupervised\/}
setting. Specifically, can a machine learn to transform a table into a chair or vice versa, in a natural way, when 
it has only seen a set of tables and a set of chairs, without any pairings between the two sets? 


\begin{figure*}[t!]
	\centering
	\includegraphics[width=\linewidth]{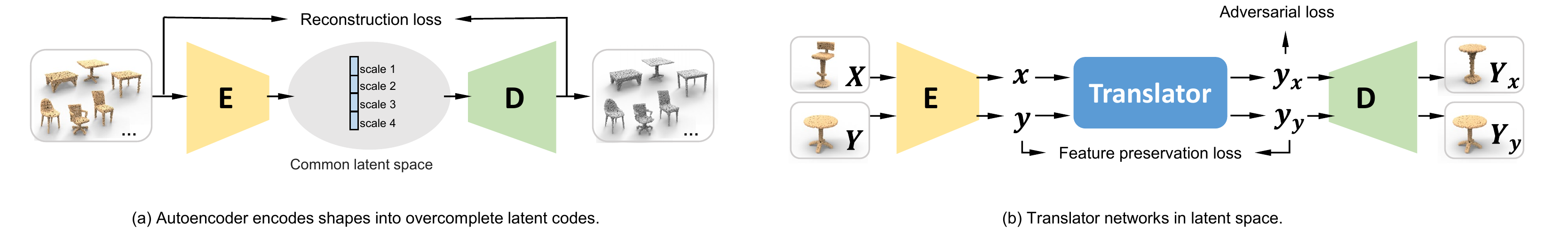}
	\caption{Overview of our network architecture, which consists of an autoencoder (a) to encode shapes from two input domains into a common latent space which is overcomplete, and a GAN-based translator network (b) designed with an adversarial loss and a loss to enforce feature preservation.}
	\label{fig:overview}
\end{figure*}

In recent years, the intriguing {\em unpaired domain translation\/} problem has drawn much interest in computer 
vision and computer graphics, e.g., the domain transfer network (DTN)~\cite{taigman2017unsupervised}, CycleGAN~\cite{zhu2017unpaired}, DualGAN~\cite{yi2017dualgan}, MUNIT~\cite{MUNIT18}, among others~\cite{UNIT17,hoshen2018NAM,hoffman2018cycada,almahairi2018augmented,gao2018automatic}.
However, most success on unpaired image-to-image translation has been achieved only on transforming 
or transferring {\em stylistic\/} image features, not shapes. A symbolic example is the CycleGAN-based cat-to-dog 
transfiguration which sees the network only able to make minimal changes to the cat/dog shapes~\cite{zhu2017unpaired}. 
The recently developed P2P-NET~\cite{yin2018p2p} is able to learn general-purpose shape transforms via point 
displacements. While significant shape changes, e.g., skeleton-to-shape or incomplete-to-complete scans, are 
possible, the training of P2P-NET is {\em supervised\/} and requires {\em paired\/} shapes from two domains. 

In this paper, we develop a deep neural network aimed at learning {\em general-purpose\/} shape transforms from 
{\em unpaired domains\/}. The network is trained on two sets of shapes, e.g., tables and chairs or different letters, 
each represented using a point cloud. There is neither a pairing between shapes in the two domains to guide the shape translation 
nor any point-wise correspondence between any shapes. Once
trained, the network takes a point-set shape from one domain and transforms it into the other.

Without any point correspondence between the source and target shapes for the transform, one
of the challenges is how to properly ``normalize'' the shapes, relating them so as to facilitate their translation. To
this end, we perform shape translation in a {\em common latent\/} space shared by the source and target domains,
rather than on the point-set shapes directly. The latent space is obtained by an {\em autoencoder\/} trained prior 
to shape transform; see Figure~\ref{fig:overview}(a).

More importantly, a proper shape transform from chairs to tables should not translate a given chair to any table,
but to a table that is clearly from that particular input chair. Hence, some features of the chair that are also common to tables
should be preserved during a chair-table translation while the other features can be altered. This poses a key 
challenge: what features are to be preserved/altered is unknown --- it must depend on the given shape domains and 
our network must learn it without supervision. To this end, our network is designed with two novel features to address this challenge:
\begin{itemize}
\item Our autoencoder encodes shape features at {\em multiple scales\/}, which is common in convolutional
neural networks (CNNs). However, unlike conventional methods which aggregate the multi-scale featuers, e.g., in 
PointNET++~\cite{qi_nips2017}, we {\em concatenate\/} the multi-scale features to produce a latent code 
\rz{which is ``{\em overcomplete\/}''. Specifically, the input shape can be reconstructed using only parts of the 
(overcomplete) code corresponding to different feature scales; see Figure~\ref{fig:ae_examples}.}

\rz{Our intuition is that performing shape transforms in the latent space formed by such overcomplete codes, where multi-scale
feature codes are {\em separated\/}, would 
facilitate an {\em implicit disentangling\/} of the preserved and altered shape features --- oftentimes, these features 
are found at different scales.} 


\item In addition, our chair-to-table translator is not only trained to turn a chair code to a table code, but 
also trained to turn a table code to the {\em same\/} table code, as shown in Figure~\ref{fig:overview}(b). 
Our motivation for the second translator loss, which we refer to as the {\em feature preservation loss\/}, is that 
it would help the translator preserve table features (in an input chair code) during chair-to-table translation.
\end{itemize}
Figure~\ref{fig:overview} shows an overview of our network architecture, with shape translation operating in a latent space 
produced by an overcomplete autoencoder. The translator network itself is built on the basic framework of generative
adversarial networks (GANs), guided by an adversarial loss and the feature preservation loss. We call our overall network a 
{\em latent overcomplete GAN\/}, or LOGAN for short, to signify the use of GANs for
shape-to-shape translation in a common, overcomplete, latent space. 
It is also possible to train a dual pair of translators between the source and target domains, reinforcing the results with an additional cyclic loss~\cite{zhu2017unpaired}. 

\begin{figure}[t!]
	\centering
	\includegraphics[width=0.9\linewidth]{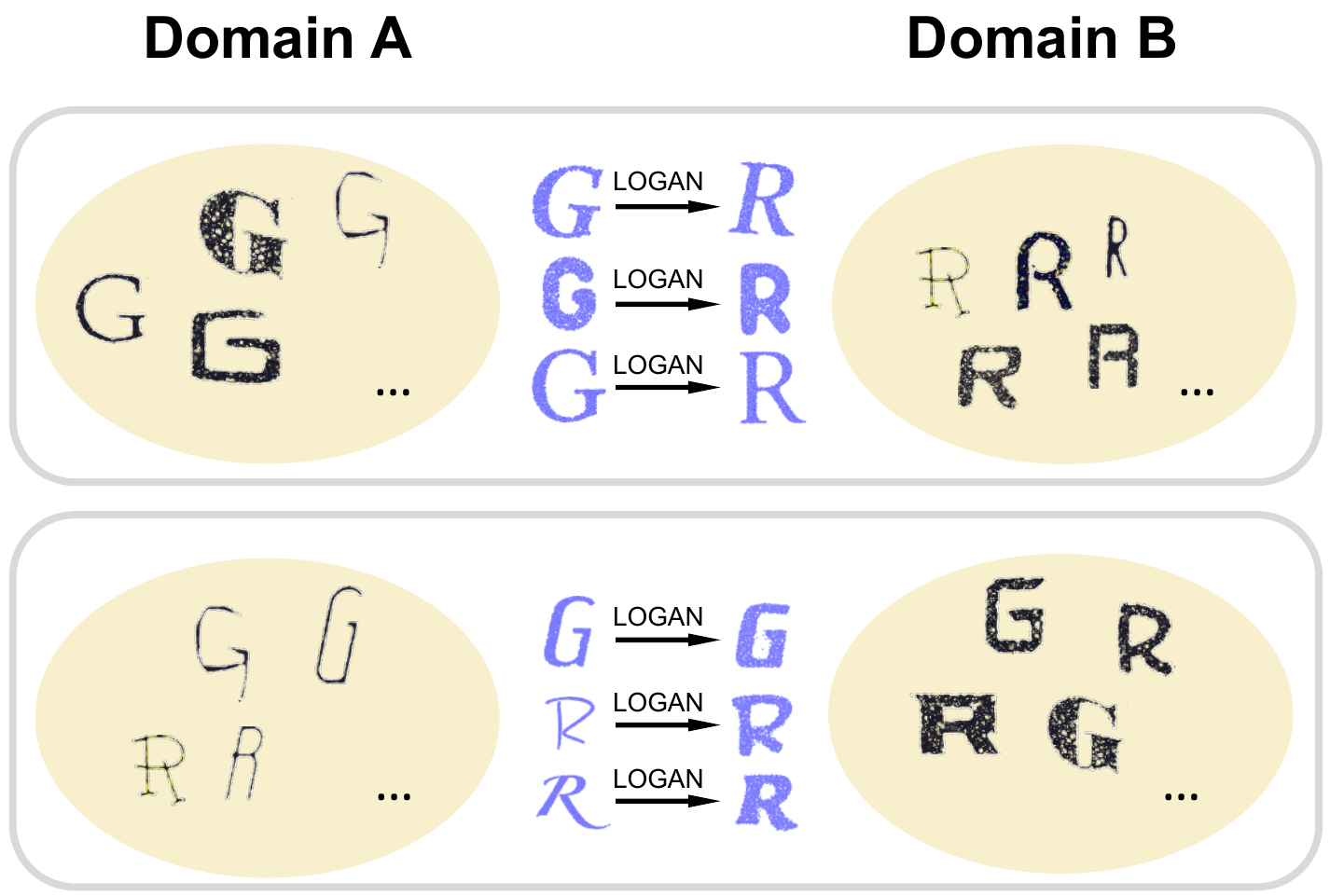}
	\caption{Depending solely on the (unpaired) input \rz{training} domains, our network LOGAN can learn both content transfer (top row: from letters $G$ to $R$, in varying font styles) and style 
transfer (bottom row: from thin to thick font strokes), without any change to the network architecture. In this example, 2D letters are represented by dense clouds of 2D points.}
	\label{fig:fontdual}
\end{figure}

Overall, LOGAN makes the following key contributions:
\begin{itemize}
\item To the best of our knowledge, it is the first deep model trained for general-purpose, unpaired shape-to-shape translation.
\item It is the first translation network that can perform {\em both\/} content and style transfers between shapes (see Figure~\ref{fig:fontdual}), 
without changing the network's architecture or any of its hyper-parameters. Owing to the overcomplete, multi-scale representations it learns, LOGAN adapts its feature preservation solely based on the two 
input domains for training.
\item It puts forth the interesting concept of implicit feature disentanglement, enabled by the overcomplete representation, which may hold potential in other application settings.
\end{itemize}


We conduct ablation studies to validate each of our key network designs: the autoencoder, the
multi-scale and overcomplete latent codes, as well as the feature preservation loss. 
We demonstrate superior capabilities in unpaired shape transforms on a variety of examples over baselines and state-of-the-art 
approaches. We show that LOGAN is able to learn what shape features to preserve during shape transforms, either 
local or non-local, whether content or style, etc., depending solely on the input domain pairs; see Figure~\ref{fig:teaser}.

\section{Related work}
\label{sec:related}

\begin{figure*}[!t]
	\centering
	\includegraphics[width=\linewidth]{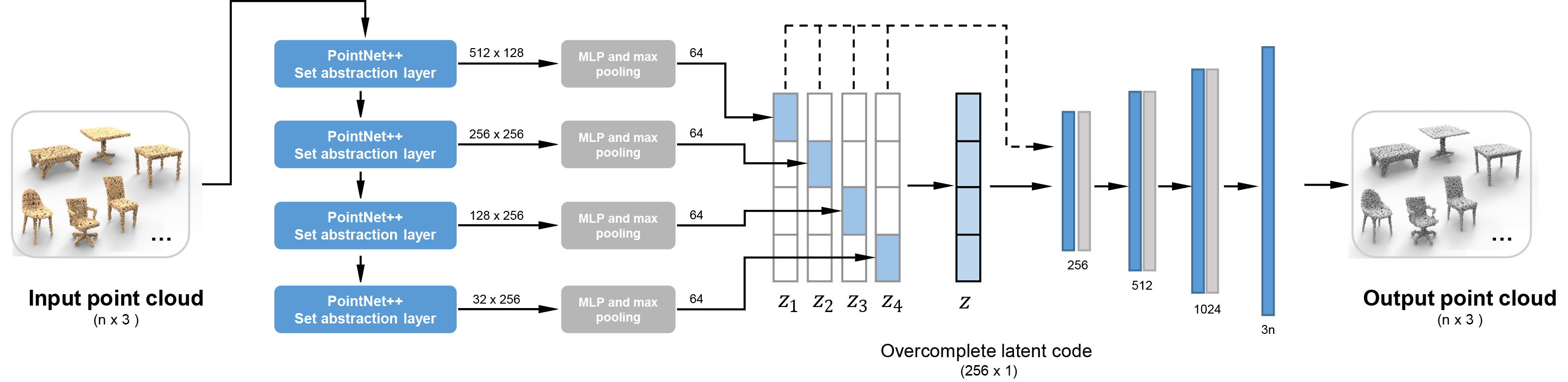}
	\caption{Architecture of our multi-scale, overcomplete autoencoder. We use the set abstraction layers of PointNet++~\cite{qi_nips2017} to produce point features in different scales and aggregate them into four sub-vectors: $z_1$, $z_2$, $z_3$, and $z_4$. The four sub-vectors are padded with zeros and summed up into a single 256-dimensional latent vector $z$ that is overcomplete; the $z$ vector can also be seen as a {\em concatenation\/} of the other four sub-vectors. During training, we feed all the five 256-dimensional vectors to the decoder. In the decoder, the blue bars represent fully-connected layers; grey bars represent ReLU layers.}
	\label{fig:autoencoder}
\end{figure*}

Computing image or shape transforms is a fundamental problem in visual data processing and covers a vast amount of literature.
In the classical setting for shape transforms, source shapes are deformed into target shapes anchored on corresponding points 
or parts. The key is how to formulate and compute deformation energies to ensure detail 
preservation~\cite{sorkine2004}, structure preservation~\cite{mitra2013survey}, or topology variation~\cite{alhashim2014topo}. 
On the other hand, our work is related to learning general-purpose, cross-domain image/shape transforms. As such, 
we mainly cover learning-based methods from vision and graphics that are most closely related to our approach.

\vspace{-5pt}

\paragraph{Unpaired image-to-image translation}
%
A wave of notable works on unsupervised/unpaired cross-domain image translation have emerged in 2017. In DTN,
Taigman et al.~\shortcite{taigman2017unsupervised} train a GAN-based domain transfer network which enforces 
consistency of the source and generated samples under a given function $f$. For example, $f$ can capture face
identities, allowing their network to generate identity-preserving emojis from facial images.
CycleGAN~\cite{zhu2017unpaired} and DualGAN~\cite{yi2017dualgan} both train dual translators with a cyclic loss
to address the challenge of unpaired image-to-image translation. However, by performing the translations on 
images directly, based on pixel-wise losses, these methods perform well on transferring stylistic images features,
but poorly on shape transforms.


Dong et al.~\shortcite{dong2017unsupervised} train a conditional GAN to learn shared global features from two image domains 
and to synthesize plausible images in either domain from a noise vector and a domain label. To enable 
image-to-image translation, they separately train an encoder to learn a mapping from an image to its latent code, which 
would serve as the noise input to the conditional GAN. 
In UNIT, Liu et al.~\shortcite{UNIT17} assume that corresponding images from two domains can be mapped to the {\em same 
code\/} in a shared latent space. Based on this assumption, they train two GANs, coupled with weight sharing, 
to learn a joint distribution over images from two unpaired domains. By sharing weight parameters corresponding to high 
level semantics in both the encoder and decoder networks, the coupled GANs are enforced to interpret these image 
semantics in the same way.

Architecturally, there are some similarities between LOGAN and UNIT~\cite{UNIT17}. Both networks take 
inputs from two domains, map them into a latent space, and enforce some notion of ``self-reconstruction''. However, 
LOGAN does not make the shared latent space/code assumption: it is not aiming to map the two inputs into the same latent code. 
Moreover, the notion of self-reconstruction in UNIT is realized by a variational autoencoder (VAE) loss and the VAE is trained together
with the GANs. In contrast, LOGAN trains its autoencoder and translator networks separately, where the notion of self-reconstruction
is applied to latent codes in the translators, via the feature preservation loss.



\vspace{-5pt}

\paragraph{Identity loss in domain translation}
Our feature preservation loss is equivalent, in form, to the {\em identity loss\/} in the DTN of Taigman et 
al.~\shortcite{taigman2017unsupervised} for reinforcing face identity preservation during emoji generation; it
was later utilized in CycleGAN~\cite{zhu2017unpaired} as an additional regularization term for color
preservation.
In our work, by enforcing the same loss in latent space, rather than on images directly, and over the multi-scale 
overcomplete codes in LOGAN, we show that the loss can play a critical role in feature preservation for a 
variety of shape transformation tasks. The preserved features can be quite versatile and adapt to the input domain pairs.


\vspace{-5pt}

\paragraph{Disentangled representations for content generation}
%
%
Disentangled image representations have been utilized to produce many-to-many mappings so as to improve 
the diversity of unsupervised image-to-image translation~\cite{lee2018diverse,MUNIT18}. Specifically, in MUNIT,
a multi-modal extension of UNIT~\cite{UNIT17}, Huang et al.~\shortcite{MUNIT18} relax
the assumption of fully shared latent space between the two input domains by postulating that only part of the 
latent space, the {\em content\/}, can be shared whereas the other part, the {\em style\/}, 
is domain-specific. Their autoencoders are trained to encode input images into a disentangled latent code 
consisting of a content part and a style part. During image translation, a fixed content code is recombined with 
a random style code to produce diverse, style-transferred target images. Most recently, Press et 
al.~\shortcite{press2019emerging} learn disentangled codes in a similar way, but for a different kind of unsupervised
content transfer task, i.e., that of adding certain information, e.g., glasses or facial hair, to source images.

In contrast, LOGAN does not learn a disentangled shape representation explicitly. Instead, our autoencoder
learns a multi-scale representation for shapes from both input domains and explicitly assigns encodings at
different shape scales to sub-vectors of the latent codes. Specifically, codes for shapes from different domains
are {\em not\/} enforced to share any sub-codes or content subspace; the codes are merely constructed in the same
manner and they belong to a common latent space. Feature preservation during shape translation is enforced in 
the translator networks, with the expectation that our overcomplete latent representation, \rz{with feature separation},
would facilitate the disentangling of preserved and altered features. \rz{Note that in other contexts
and for other applications using CNNs, there have been works, e.g.,~\cite{eigen2014,song2016ssc,cui2016}, 
which also encode and concatenate multi-scale features.}




\vspace{-5pt}

\paragraph{Learning shape motions and transforms}
Earlier work on spatial transformer networks~\cite{jaderberg2015spatial} allows deep convolutional models to 
learn invariance to translation, scale, rotation, and more generic shape warping for improved object recognition.
Byravan and Fox~\shortcite{byravan2016se3} develop a deep neural network to learn rigid body motions
for robotic applications, while deep reinforcement learning has been employed to model controllers for
a variety of character motions and skills~\cite{peng2018deepMimic}. 
%
%
For shape transforms, Berkiten et al.~\shortcite{berkiten2017detail} present a metric learning approach for
analogy-based mesh detail transfer. In P2P-NET, Yin et al.~\shortcite{yin2018p2p} develop a point 
displacement network which learns transforms between point-set shapes from two {\em paired\/} domains. 

More closely related to our work is the VAE-CycleGAN recently developed by Gao et al.~\shortcite{gao2018automatic} 
for unpaired {\em shape deformation transfer\/}. Their network is trained on two unpaired animated mesh sequences,
e.g., animations of a camel and a horse or animations of two humans with different builds. Then, given a deformation
sample from one set, the network generates a shape belonging to the other set which possesses the same pose. 
One can view this problem as a special instance of the general shape transform problem. Specifically, it is a
pose-preserving shape transform where the source meshes (respectively, the target meshes) model different poses 
of the same shape and they all have the same mesh connectivity. LOGAN, on the other hand, is designed to be a 
general-purpose translation network for point-set shapes, where much greater geometric and topological variations
among the source or target shapes are allowed.

Technically, the VAE-CycleGAN of Gao et al.~\shortcite{gao2018automatic} encodes each input set into
a {\em separate\/} latent space and trains a CycleGAN to translate codes between the two latent spaces. 3D models can 
then be recovered from the latent codes by the VAE decoder. In contrast, LOGAN encodes shapes from both
input domains into a common latent space and performs shape translations in that space. To enable generic shape
transforms, the key challenge we address is learning what shape features to preserve during the translation.

\vspace{-5pt}

\paragraph{Deep learning for point-set shapes}
Recently, several deep neural networks, including PointNET~\cite{qi_cvpr2017}, PointNET++~\cite{qi_nips2017}, 
PCPNET~\cite{guerrero2017pcpnet}, PointCNN~\cite{li2018pointcnn}, and PCNN~\cite{atzmon2018PCNN}, have been developed 
for feature learning over point clouds. Generative models of point-set shapes~\cite{fan2016point,achlioptas2018learning} and
supervised, general-purpose point-set transforms~\cite{yin2018p2p} have also been proposed. To the best
of our knowledge, our work represents the first attempt at learning general shape transforms from unpaired domains.
While LOGAN relies on PointNET++ for its multi-scale feature encodings, it produces an overcomplete latent
code via feature concatenation rather than feature aggregation.


\section{Method}
\label{sec:method}

Given two sets of unpaired shapes $\mathcal{X}$ and $\mathcal{Y}$, our goal is to establish two mappings $\mathcal{M}_{\mathcal{X} \rightarrow \mathcal{Y}}: \mathcal{X} \mapsto \mathcal{Y}$ and $\mathcal{M}_{\mathcal{Y} \rightarrow \mathcal{X}}: \mathcal{Y} \mapsto \mathcal{X}$, to translate shapes between the two domains. The translation should be natural and intuitive, with  emphasis given to the preservation of common features. We make no special assumptions on the two domains, except that certain common features exist in both domains. Note that such features can be both local and global in nature, and hence are difficult to define or model directly. Therefore, we employ deep neural networks to implicitly learn those features.

\subsection{Overview of networks and network loss}
\label{sec:loss}

As shown in Figure~\ref{fig:overview}, our network comprises of two parts that are trained in separate steps.
First, an autoencoder is trained. The multi-scale encoder (Sec.~\ref{sec:methodAutoencoder}) $E$ takes point clouds from both domains as input, and encodes them into compact latent codes in a common latent space. The decoder $D$ decodes the latent codes back into point clouds. After training, the autoencoder produces the over-complete latent codes for the input shapes, denoted by $\mathcal{Z_X}$ and $\mathcal{Z_Y}$, where
\revise{
$\mathcal{Z_X}=\{E(X) \:|\: X \in \mathcal{X} \}$ and $\mathcal{Z_Y}=\{E(Y)  \:|\: Y \in \mathcal{Y}\}$.
}

The second part of our network is a latent code translator network that transforms between $\mathcal{Z_X}$ and $\mathcal{Z_Y}$. It consists of two translators: $T_{\mathcal{X} \rightarrow \mathcal{Y}}: \mathcal{Z_X} \mapsto \mathcal{Z_Y}$ and $T_{\mathcal{Y} \rightarrow \mathcal{X}}: \mathcal{Z_Y} \mapsto \mathcal{Z_X}$. We only show $T_{\mathcal{X} \rightarrow \mathcal{Y}}$ in the figure for simplicity. The translators take latent codes in both domains as input, and treat them differently with two different loss functions. Once trained, given a shape $X \in \mathcal{X}$, its translated shape in domain $\mathcal{Y}$ is obtained by $Y_x = D(T_{\mathcal{X} \rightarrow \mathcal{Y}}(E(X)))$.

We use three loss terms for the translator network to create a natural and feature-preserving mapping:
\begin{itemize}
\item {\em Adversarial loss\/}:
We take $T_{\mathcal{X} \rightarrow \mathcal{Y}}$ in Figure~\ref{fig:overview}(b) as an example. Given
$x \in \mathcal{Z_X}$, the network performs the translation and an adversarial loss is applied on the translated
latent-code. The discriminator would judge the output codes from the ground-truth codes in $\mathcal{Z_Y}$, 
to ensure that the distribution of the translator outputs matches the target distribution of $\mathcal{Z_Y}$.
\item {\em Feature preservation loss\/}:
Given $y \in \mathcal{Z_Y}$, since this network serves only $\mathcal{X} \rightarrow \mathcal{Y}$ transfer, the
output still falls in $\mathcal{Z_Y}$, and we use a feature preservation loss (identity loss) to enforce the output of
the network to be similar to the original input $y$. Feature preservation loss is the key for our network to learn
meaningful mappings, since it encourages the translator to keep most portions of the code intact and only changes
the parts that are really important for the domain translation.
\item {\em Cycle-consistency loss\/}:
The two loss terms above play critical roles in our translator network and already allow the generation of satisfactory 
results. However, we may introduce a third loss term, the cycle-consistency loss or simply, the cycle loss, to further 
regularize the translation results.

The cycle loss pushes each shape to reconstruct itself after being translated to the opposite domain and then translated 
back to its original domain. The term encourages the mappings 
$T_{\mathcal{X} \rightarrow \mathcal{Y}}$ and $T_{\mathcal{Y} \rightarrow \mathcal{X}}$ to be one-to-one. It further 
reduces the possibility that shapes in one domain only map to a handful of shapes in the opposite domain, i.e., mode 
collapse.
\end{itemize}

Note that our translator network still maintains a cross-domain similarity, which results from a joint effort of the 
feature preservation and adversarial losses --- the former loss forces the result to be similar to the source shape, 
while the latter loss ensures plausibility in the target domain.
The cycle loss only serves a supportive role in our network. Without cycle-consistency loss,
our network is still able to suppress mode collapse. The multi-scale encoder pushes the codes from both domains 
to share the common space in different scales, bringing large distribution overlap and making it hard to collapse in 
$\mathcal{Z_X} \rightarrow \mathcal{Z_Y}$ without collapsing in $\mathcal{Z_Y} \rightarrow \mathcal{Z_Y}$, and 
the latter is unlikely to happen due to the feature preservation loss. In our experiments, cycle loss becomes prominent 
when the size of the dataset is very small. Details of the translator network and the loss functions can be found in 
Sec.~\ref{sec:methodTranslator}.

\subsection{Multi-scale overcomplete autoencoder}
\label{sec:methodAutoencoder}

\begin{figure}[!t]
	\centering
	\includegraphics[width=0.95\linewidth]{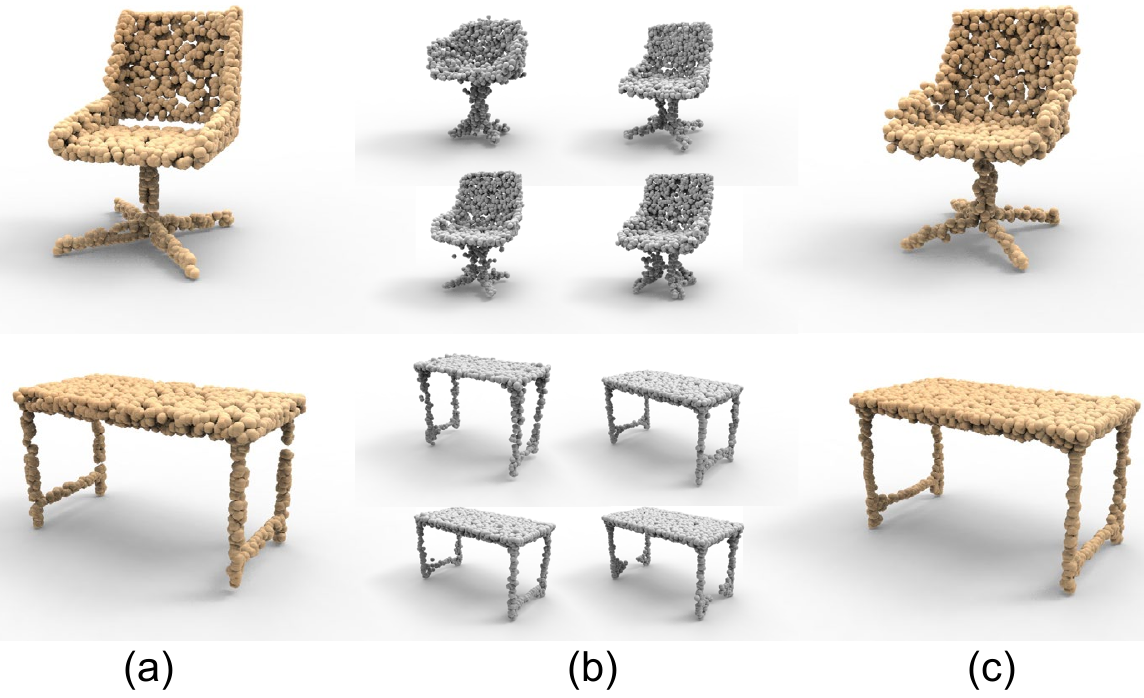}
	\caption{Our autoencoder encodes each test point cloud (a) into 5 latent vectors ($z$, $z_1,\ldots,z_4$), as shown in Figure~\ref{fig:autoencoder}, and decodes them back to point clouds. The decoder output (c) for the overcomplete latent code $z$ exhibits better reconstruction than those from the other sub-vectors (b). \rz{Yet, all the reconstructions in (b) and (c) resemble the input shapes well, demonstrating the overcompleteness of our latent codes.}}
	\label{fig:ae_examples}
\end{figure}

Our multi-scale autoencoder is depicted in Figure~\ref{fig:autoencoder}. The input to our encoder is a set of $n$ points.
It passes through four set abstraction layers of PointNet++ ~\cite{qi_nips2017} with increasing sampling radius. The output point features from each of the four layers are further processed by an MLP and a max-pooling layer to form a single 64-dimensional sub-vector. We pad the 4 sub-vectors $z_1$, $z_2$, $z_3$, $z_4$ from the 4 branches with zeros to make them 256-dimensional vectors, and sum them up to get an overcomplete 256 dimensional latent vector $z$. The detailed network structure can be found in the supplementary material.


During training, we feed the padded sub-vectors and the overcomplete latent vector to the same decoder. Similar to ~\cite{achlioptas2018learning}, our decoder is a multilayer perceptron (MLP) with 4 fully-connected layers. Each of the first 3 fully-connected layers are followed by a ReLU layer for non-linear activation. The last fully-connected layer outputs an $n \times 3$ point cloud for each input 256-dimensional vector. As shown in Figure~\ref{fig:ae_examples}, our decoder is able to reconstruct point clouds from the 5 vectors simultaneously. The quality of reconstruction from $z$ is higher than the 4 sub-vectors as it contains  most information. The loss function of the autoencoder considers all the 5 point clouds reconstructed from the 5 vectors:
\begin{equation}\label{eq:AE_loss}
{L}_{\mathrm{AE}} = {L}_{z}^{\mathrm{rec}} + \lambda_1 \sum_{i=1}^{4}{{L}_{z_i}^{\mathrm{rec}}},
\end{equation}
where $\lambda_1$ is a scalar weight set to 0.1 by default. ${L}_{z}^{\mathrm{rec}}$
and ${L}_{z_i}^{\mathrm{rec}}$ denote reconstruction losses for code $z$ and $z_i$. \revise{When training the autoencoder, the input point cloud contains 2,048 points. Our decoder produces 2,048 points from each of the 5 latent vectors.} \rz{Note that we choose the Earth Mover's Distance (EMD)~\cite{rubner2000earth} to define the reconstruction losses for all the 5 latent vectors, since EMD has been found to produce less noisy outputs compared to Chamfer Distance~\cite{fan2016point}.}

Note that our current choices of the latent code length (256) and number of scales (4) are both empirical. We tested autoencoding using shorter/longer codes
as well as scale counts from two to six. Generally, short codes do not allow the multi-scale shape features to be well encoded and using too few scales compromises
the translator's ability to disentangle and preserve the right features during cross-domain translation.
On the other hand, employing latent codes longer than 256 or increasing the number of scales beyond 4 would only introduce extra redundancy in the latent space,
which did not improve translation results in our experiments.

\subsection{Feature-preserving shape transform}
\label{sec:methodTranslator}

\begin{figure}[!t]
	\centering
	\includegraphics[width=\linewidth]{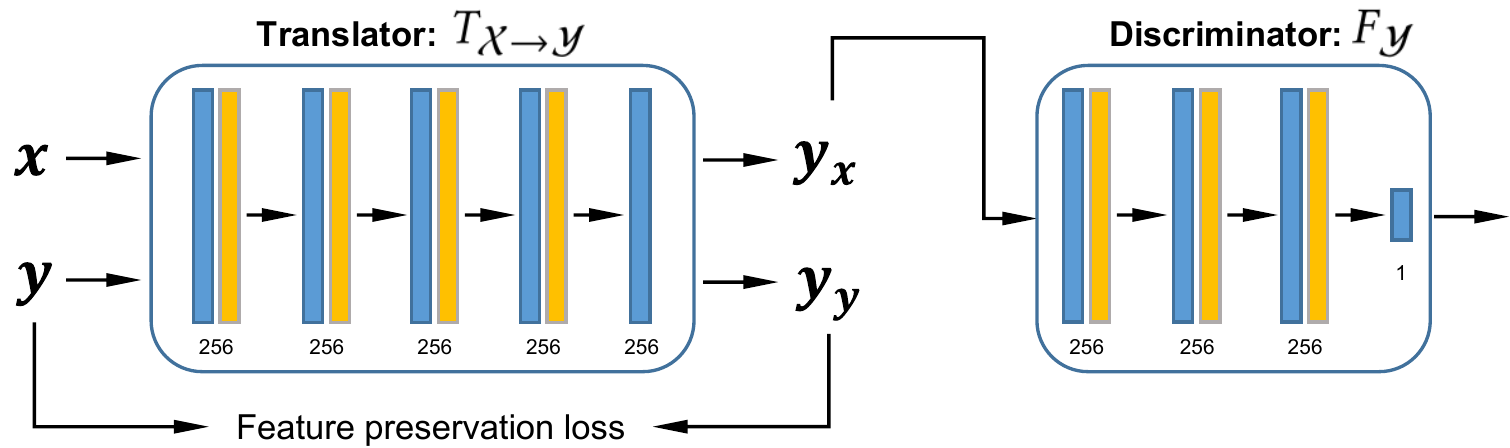}
	\caption{Architecture of our translator network.  The blue bars represent fully-connected layers; orange bars represent BN-ReLU layers. }
	\label{fig:translator_archi}
\end{figure}

Our translators $T_{\mathcal{X} \rightarrow \mathcal{Y}}$ and $T_{\mathcal{Y} \rightarrow \mathcal{X}}$ work in the common latent space.
Similar to the decoder, they are implemented as MLPs with 5 fully-connected (FC) layers, as shown in Figure~\ref{fig:translator_archi}. The detailed network structure can be found in the supplementary material.

The two discriminators $F_\mathcal{X}$ and $F_\mathcal{Y}$ work in the latent space as well. They are implemented as MLPs with 3 FC hidden layers where each of them are followed by a BN layer and a ReLU layer, as shown in  Figure~\ref{fig:translator_archi}.
We adopt WGAN ~\cite{arjovsky2017wasserstein} in our implementation. To that end, we directly take the result of the output layer without sigmoid activation.

In the common latent space, the translators and discriminators work directly over the over-complete latent code, $x \in \mathcal{Z_X}$ and $y \in \mathcal{Z_Y}$,  as they already contain the information of the sub-vectors.
For simplification, in the following part of this section, we will only explain the loss function in details for $T_{\mathcal{X} \rightarrow \mathcal{Y}}: \mathcal{Z_X} \mapsto \mathcal{Z_Y}$. The opposite direction can be derived directly by swapping $x$ and $y$ in the equations. The loss function  for $T_{\mathcal{X} \rightarrow \mathcal{Y}}$ is,
\begin{equation}
\begin{aligned}
{L}_{\mathcal{X} \rightarrow \mathcal{Y}} =  {L}^{\mathrm{WGAN}}_{\mathcal{X} \rightarrow \mathcal{Y}} + \alpha {L}^{\mathrm{FP}}_{\mathcal{X} \rightarrow \mathcal{Y}}
\end{aligned}
\end{equation}
where $\alpha$ is a scalar weight set to 20 by default.
Similar to WGAN ~\cite{gulrajani2017improved}, the adversarial loss for $T_{\mathcal{X} \rightarrow \mathcal{Y}}$ is defined as:
\begin{equation}
\begin{aligned}
{L}^{\mathrm{WGAN}}_{\mathcal{X} \rightarrow \mathcal{Y}} =  \mathbb{E}_{y \sim \mathbb{P} (\mathcal{Z_Y})} [{F} _\mathcal{Y}(y)]
- \mathbb{E}_{x \sim \mathbb{P}(\mathcal{Z_X})} [{F}_\mathcal{Y}( y_x )] + \lambda_2 L_{GP}
\end{aligned}
\end{equation}
where $y_x = T_{\mathcal{X} \rightarrow \mathcal{Y}}(x)$ is the output of translation  for $x$. $L_{GP}$ is the gradient penalty term introduced by ~\cite{gulrajani2017improved} for regularization. $\lambda_2$ is a scalar weight set to 10 by default.
During training, our discriminator ${F} _\mathcal{Y}$ aims to maximize the adversarial loss, while our translator $ T_{\mathcal{X} \rightarrow \mathcal{Y}}$ aims to  minimize it.

As shown in Figure~\ref{fig:translator_archi}, our feature preservation loss is defined in the latent space. For translator $T_{\mathcal{X} \rightarrow \mathcal{Y}}$, it is defined  as the $L1$ distance between the input vector $y \in \mathcal{Z_Y}$ and its translated output vector $y_y = T_{\mathcal{X} \rightarrow \mathcal{Y}} (y)$ :
\begin{equation}
{L}^{\mathrm{FP}}_{\mathcal{X} \rightarrow \mathcal{Y}}  = \mathbb{E}_{y \sim \mathbb{P}(\mathcal{Z_Y})} [ \lVert y - T_{\mathcal{X} \rightarrow \mathcal{Y}} (y)\rVert _1 ]
\end{equation}

Training our translators with the loss function ${L}_{\mathcal{X} \rightarrow \mathcal{Y}}$ or ${L}_{\mathcal{Y} \rightarrow \mathcal{X}}$  is able to produce reasonable result as shown in Figure~\ref{fig:comp_chairtable} (g). However, having a cycle-consistency term could further improve the result by encouraging one-to-one mapping between the two input domains.
For latent code $x \in \mathcal{Z_X}$, applying $T_{\mathcal{X} \rightarrow \mathcal{Y}}$ followed by $T_{\mathcal{Y} \rightarrow \mathcal{X}}$ should produce a code similar to itself: $T_{\mathcal{Y} \rightarrow \mathcal{X}} ( T_{\mathcal{X} \rightarrow \mathcal{Y}} (x) ) \approx x$. Similarly for $y \in \mathcal{Z_Y}$ we have: $T_{\mathcal{X} \rightarrow \mathcal{Y}} ( T_{\mathcal{Y} \rightarrow \mathcal{X}} (y) ) \approx y$. Thus, the cycle-consistency loss term is defined as,
\begin{equation}
\begin{aligned}
{L}_{\mathrm{Cycle}}  & = \mathbb{E}_{x \sim \mathbb{P}(\mathcal{Z_X})} [ \lVert T_{\mathcal{Y} \rightarrow \mathcal{X}} ( T_{\mathcal{X} \rightarrow \mathcal{Y}} (x) ) - x \rVert _1 ] \\
& + \mathbb{E}_{y \sim \mathbb{P}(\mathcal{Z_Y})} [ \lVert T_{\mathcal{X} \rightarrow \mathcal{Y}} ( T_{\mathcal{Y} \rightarrow \mathcal{X}} (y) ) - y \rVert _1 ]
\end{aligned}
\end{equation}

The overall loss function is defined as:
\begin{equation}
\begin{aligned}
{L}_\mathrm{Overall} =  {L}_{\mathcal{X} \rightarrow \mathcal{Y}} + {L}_{\mathcal{Y} \rightarrow \mathcal{X}} + \beta {L}_{\mathrm{Cycle}}
\end{aligned}
\end{equation}
where $\beta$  is a  scalar weight set to 20 by default.
With the overall loss function, the goal of training the translators is to solve:
\begin{equation}
T^*_{\mathcal{X} \rightarrow \mathcal{Y}}, T^*_{\mathcal{Y} \rightarrow \mathcal{X}} = \arg \min_{T} \max_{F} {L}_{\mathrm{Overall}}
\end{equation}
where $F$ denotes $\{ F_\mathcal{X},  F_\mathcal{Y} \}$, $T$ denotes $ \{  T_{\mathcal{X} \rightarrow \mathcal{Y} }, T_{\mathcal{Y} \rightarrow \mathcal{X}} \}$.
In Sec.~\ref{sec:results}, we perform an ablation study to show that all loss terms play active roles in achieving high-quality results.

\subsection{Training details and point upsampling }
\label{sec:methodTraining}

We train the AE and the translator networks separately in two steps, since an insufficiently trained autoencoder can misguide the translators into poor local minima.
In our experiments, we train the autoencoder for 400 epochs with an Adam optimizer (learning rate = 0.0005, batch size = 32). After that, we train the translator networks with Adam and the training schema of ~\cite{gulrajani2017improved} for 600 epochs. We set the number of discriminator iterations per generator iteration to two. The batch size we set for training the translator networks is 128. \rz{The learning rate starts at 0.002 and is halved after every 100 training epochs, until reaching 5e-4.}
Assuming each of the datasets of two domains contains 5000 shapes, the training of autoencoder takes about 20 hours and the training of two translators takes about 10 mins on a NVIDIA Titan Xp GPU. 

\begin{figure}[!t]
\centering
\includegraphics[width=0.9\linewidth]{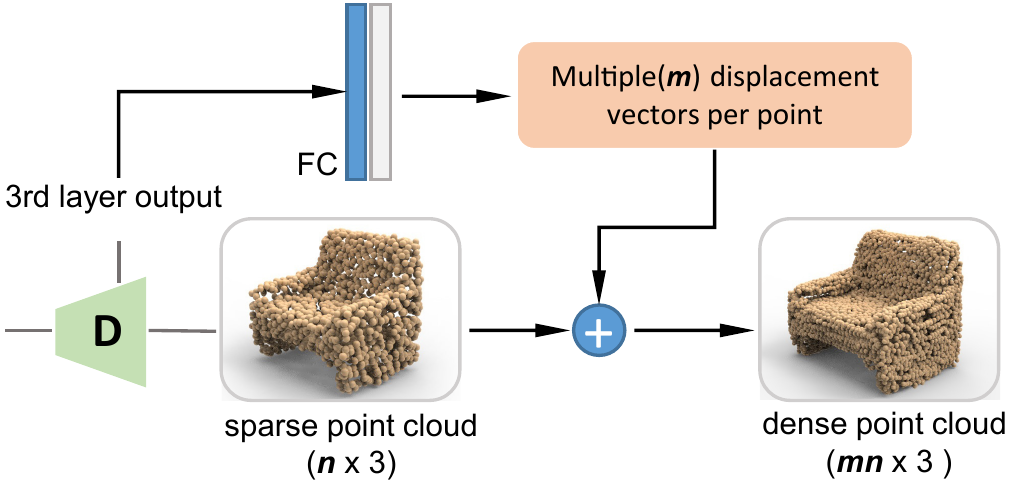}
\caption{Architecture of the upsampling layer of our network after shape translation. We predict $m$ local displacement vectors for each of the $n$ points in the sparse point cloud, which results in a dense set of $mn$ points.}
\label{fig:upsample_archi}
\end{figure}

By default, we train LOGAN on point clouds each of which contains 2,048 points and the networks output point clouds at the same resolution. However, we allow the addition of an {\em upsampling layer\/} to our network to produce higher-resolution outputs when desired. Specifically, we attach a single upsampling layer to the {\em second-to-last\/} layer of the trained decoder.  We train the upsampling layer independently after the shape translation has been complete, so it would not affect the translation results.  As shown in Figure~\ref{fig:upsample_archi},  the upsampling layer recombines the neural outputs from the second-to-last layer and predicts $m$ local displacement vectors for each point in the output point cloud. As a result, it splits each point into $m$ points and turns the sparse point cloud of size $n$ into a dense point cloud of size $mn$. The upsampling layer is a fully-connected layer which is followed by a sigmoid function. We scale the output of the sigmoid function to make it lie in the interval $[-0.05, 0.05]$.  

To train the upsampling layer, we sample 16,384 points from  each training shape as the ground truth (GT). The loss function is defined as the distance between the dense point cloud of size $mn$  and the dense GT point cloud.  
\revise{To reduce computation and memory costs, we randomly select 4,096 points from each of the two dense point clouds, and compute the EMD between the downsampled point clouds as an approximation of the EMD between the dense point clouds.}
Note that sometimes one of the two domains does not require upsampling (e.g., meso-skeletons). In that case, we train the upsampling layer with only the dataset of the other domain.
The upsampling layer is trained independently for 80 epochs, with an Adam optimizer (learning rate = 0.0005, batch size = 32).  Assuming that each of the two domains contains 5,000 shapes, training the upsampler takes about 5 hours  on a NVIDIA Titan Xp GPU. 

\section{Results and evaluation}
\label{sec:results}

\begin{figure*}[t!]
	\centering
	\includegraphics[width=0.9\linewidth]{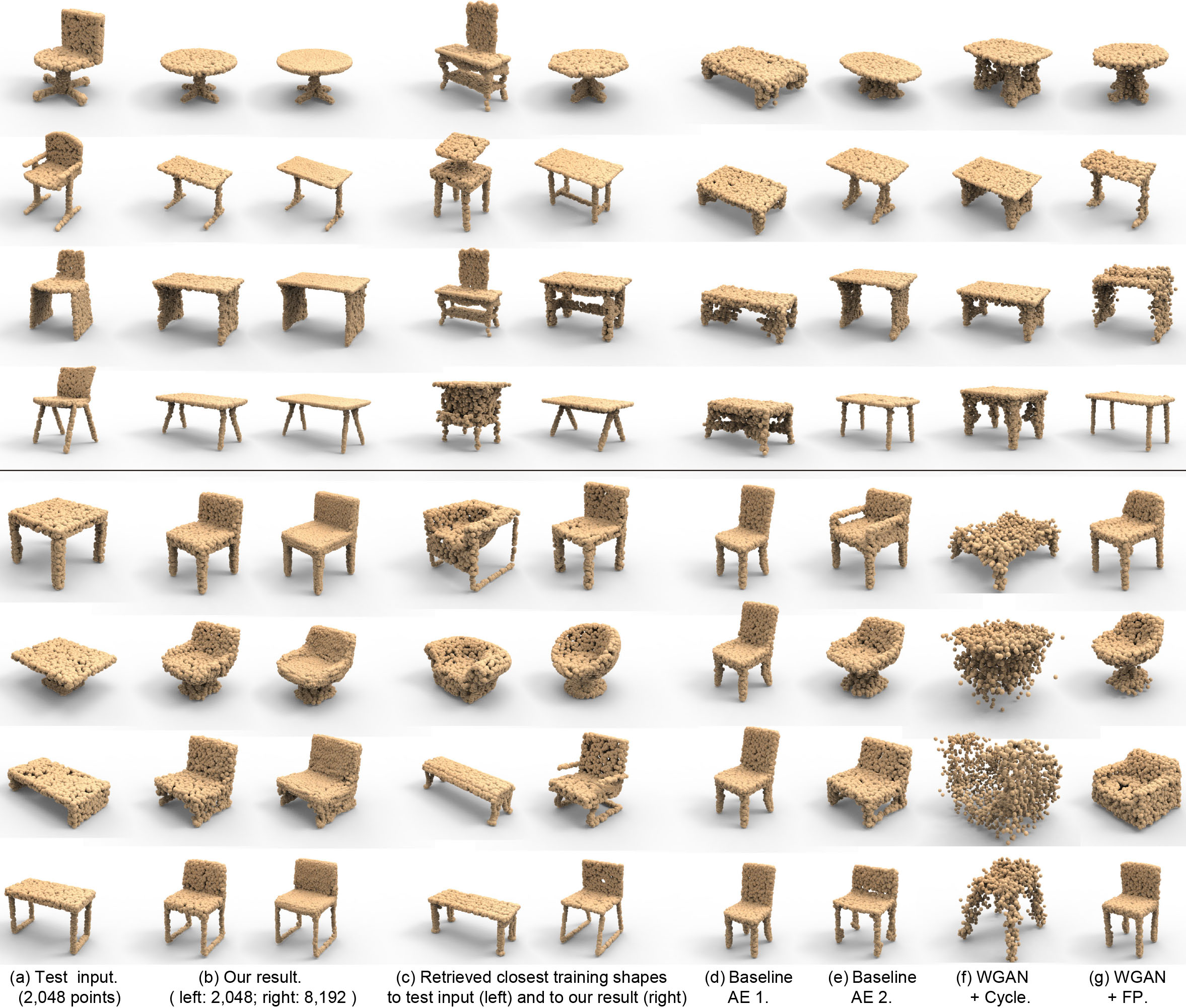}
	\caption{Comparing chair-table translation results using different network configurations. Top four rows: chair $\rightarrow$ table. Rest: table  $\rightarrow$  chair. 
	(a) Test input. 
	(b) LOGAN results with and without upsampling. 
	(c) Retrieved training shapes from the {\em target\/} domain which are closest to the test input (left) and to our translator output (right). The retrieval was based on EMD between point clouds at 2,048 point resolution. Note that the chair dataset from ShapeNet has some benches mixed in, which are retrieved as ``tables.'' 
	(d) Baseline AE 1 as autoencoder $+$ our translator network. 
	(e) Baseline AE 2 ($\lambda_1=0$) $+$ our translator network. 
	(f) Our autoencoder ($\lambda_1=0.1$) + WGAN \& Cycle loss. 
	(g) Our autoencoder ($\lambda_1=0.1$) + WGAN \& feature preservation (FP) loss.
	}
	\label{fig:comp_chairtable}
\end{figure*}

%

To demonstrate the capability of LOGAN in learning unpaired shape transforms, we conduct experiments including ablation studies and comparisons to baselines and state-of-the-art techniques. Throughout the experiments, specifically for all results shown in Sections~\ref{subsec:result_baseline},  \ref{subsec:style_content},  \ref{subsec:imp_dis} and \ref{subsec:comp_p2p}, LOGAN was trained with the {\em same\/} default network settings as described in Section~\ref{sec:method} and the supplementary material; there is no hyperparameter or architecture tuning for any specific input datasets. The only tuning was applied when training LOGAN on the small datasets of~\cite{gao2018automatic}, to avoid overfitting.
All visual results are presented without any post-processing.


\subsection{Shape transform results and ablation studies}
\label{subsec:result_baseline}

The first domain pair we tested LOGAN on is the chair and table datasets from ShapeNet Core~\cite{chang2015_shapeNet}, which contains mesh models. The chair dataset consists of 4,768 training shapes and 2,010 test shapes, while the table dataset has 5,933 training and 2,526 test shapes. We normalize each chair/table mesh to make the diagonal of its bounding box equal to unit length and sample the normalized mesh uniformly at random to obtain 2,048 points for our point-set shape representation.  
If upsampling is required, we sample from each training mesh another set of 16,384 points with Poisson disk sampling~\cite{corsini2012efficient} to form the training dataset for the upsampler. But note that the autoencoder and the translator networks are always trained by point clouds of size 2,048.
%

\paragraph{Comparing autoencoding.}
With the chair-table domain pair, we first compare our autoencoder, which produces multi-scale and overcomplete latent codes, with two baseline alternatives:
\begin{itemize}
\item In Baseline AE 1, we apply the original PointNet++, as described in~\cite{qi_nips2017}, as the encoder to produce latent vectors of length 256 (the same as in our autoencoder) and use the same decoder as described in Section ~\ref{sec:methodAutoencoder}. With this alternative, there is no separate encoding of mulit-scale features (into sub-vectors as in our case) to produce an overcomplete latent code; features from all scales are aggregated. 
\item In Baseline AE 2, we set $\lambda _1 = 0$ in the loss function (\ref{eq:AE_loss}) of our autoencoder. With this alternative, the autoencoder still accounts for shape features from all scales (via the vector $z$), but the impact of each sub-vector (one of the $z_i$'s, $i=1,\ldots,4$) for a specific feature scale is diminished.
%
%
\end{itemize}

An examination on reconstruction errors of the three autoencoders, based on the Earth Mover's Distance (EMD), reveals that our autoencoder may not be the best at reconstructing shapes. However, the main design goal of our autoencoder is to facilitate shape translation between unpaired domains, not accurate self-reconstruction. 

In Figure~\ref{fig:comp_chairtable} (b, d, e), we show that with the same translator network but operating in different latent spaces, our autoencoder leads to the best cross-domain transforms, compared to the two baselines. 

With Baseline AE 1, the translator is unable to preserve input features and can suffer from mode collapse. With a multi-scale overcomplete code $z$, Baseline AE 2 clearly improves results, but it can still miss input features that should be preserved, e.g., more global features such as the roundness at the top (row 1), the oblique angles at the legs (row 4), and more local features such as the bottom slat between the legs (row 8); it could also add erroneous features such as armrests (row 5) and extra holes (row 7). 

In contrast, with a more overcomplete and multi-scale encoding into the latent space by using all five vectors ($z, z_1, \ldots, z_4$), our default autoencoder produces the most natural table-chair translations. This is likely attributed to a better disentangling of the preserved and altered features in the latent codes.

\rz{Note that the results shown in Figure~\ref{fig:comp_chairtable} were casually picked as representative examples to demonstrate the capability of our network. Larger sets of randomly selected outputs can be found in the supplementary material and they reveal the same trend: our method consistently outperforms the baselines. }

\paragraph{Comparison to retrieval results.}
As a sanity check, we show that our network indeed {\em generates\/} shape transforms; it does not simply retrieve a training shape from the target domain. In Figure~\ref{fig:comp_chairtable}, \revise{column (c)}, we show retrieved training shapes from the target domain that are the closest (based on EMD) to the test input and to our translator output. It is quite evident that in general, these training shapes are far from being similar to outputs from our translator, with a couple of exceptions shown in rows 4 and 5. In these rare cases, it so happens that there are similar shapes in the target training set.

\paragraph{Joint embedding of latent codes.}

\begin{figure}[t!]
	\centering
	\includegraphics[width=0.9\linewidth]{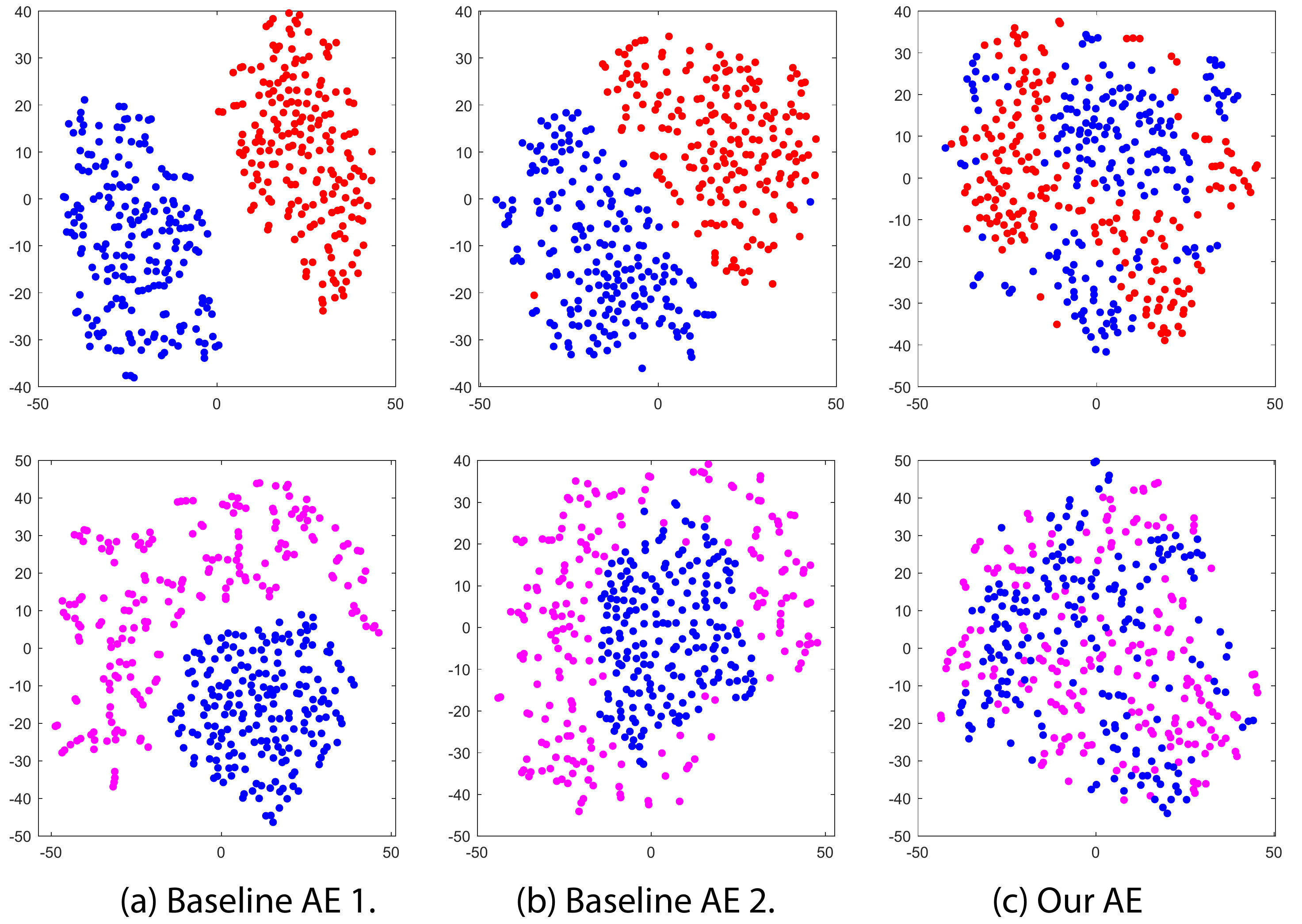}
	\caption{Visualizing joint embeddings of table and chair latent codes produced by three AEs. Top row: red = chair, blue = table, {\em before\/} translation. Bottom: magenta = chair {\em after\/} translation; blue = original table. Our default AE brings the chairs and tables closer together in common latent space.}
	\label{fig:embedding_viz}
\end{figure}

Figure~\ref{fig:embedding_viz} visualizes the common latent spaces constructed by the three AEs by jointly embedding the chair and table latent codes. For each domain, we standardize every latent dimension by moving the mean to 0 and scaling the values to a standard deviation of 1.0. We discretize the values in each dimension by multiplying it by 3 and rounding it to an integer.  Finally, we measure distances between all the latent codes using Hamming distance and embed the codes into 2D space via t-SNE.

We can observe that, compared to the two baselines, our default AE brings the chairs and tables closer together in the latent space, before the translation, effectively ``tangling" together the distributions of the generated latent codes to better discover their common features. During translation, the overcomplete codes facilitate an implicit disentanglement of the preserved and altered features. After chair$\rightarrow$table translation, the chair codes are closer to the tables in all three cases, but our default network clearly produces a better coverage of the target (table) domain. This makes the translated chair latent codes more plausible in the table domain, which can explain, in part, a superior performance for the translation task.


\paragraph{Comparing translator settings.}
%
In the second ablation study, we fix our autoencoder as presented in Figure~\ref{fig:autoencoder}, but change the translator network configuration by altering the loss function into two baseline alternatives: WGAN loss $+$ Cycle loss and WGAN loss $+$ feature preservation (FP) loss. Note that our default network LOGAN has all three losses. It is quite evident, from the visual results in Figure~\ref{fig:comp_chairtable}, that the feature preservation loss has significant positive impact on cross-domain translation, while the cycle loss provides additional regularization for improved results.

\paragraph{Part removal/insertion.}

\begin{figure}[t!]
	\centering
	\includegraphics[width=0.95\linewidth]{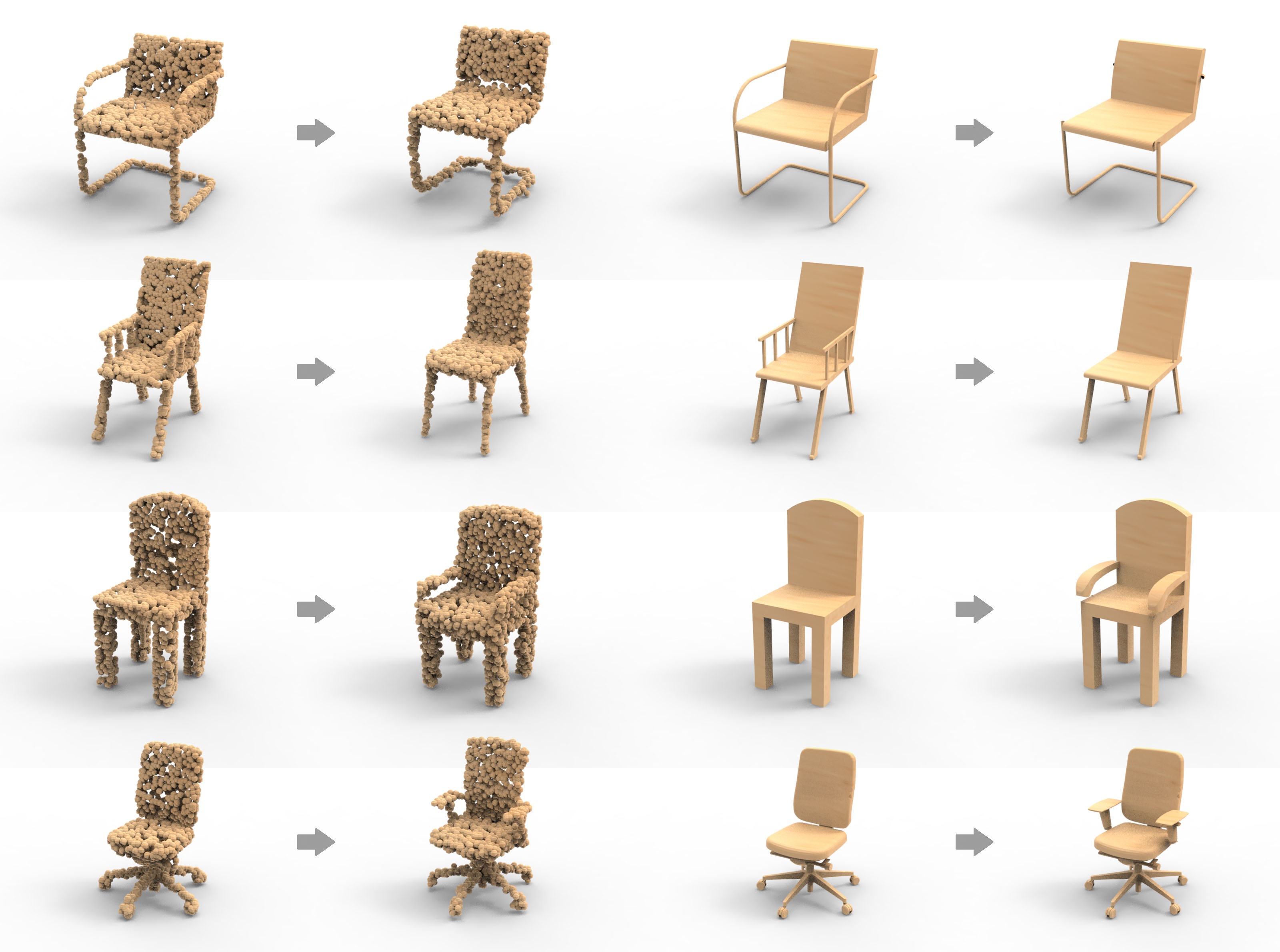}
	\caption{Unpaired shape transforms between armchairs and armless chairs. The first two rows show results of armrest removal by LOGAN, while the last two rows show insertion. On the right, we show the mesh editing results guided by the learned point cloud transforms.}
	\label{fig:armchair_armless}
\end{figure}

Chair-table translations mainly involve transforms between local shape structures. As another example, we show that LOGAN is able to learn to remove/add armrests for the chair dataset; see Figure~\ref{fig:armchair_armless}. The dataset split was obtained from the chairs in ShapeNet Core by a hand-crafted classifier. It contains 2,138 armchairs and 3,572 armless chairs, where we used 80\% of the data for training and the rest for testing. The results demonstrate that our network can work effectively on part-level manipulation, as it learns which parts to alter and which parts to preserve purely based on the observation of the input shapes from the two domains, without supervised training. In addition, the insertion/removal of the armrest parts are carried out naturally.  

We show in Figure~\ref{fig:armchair_armless} (right) that we can use the learned point cloud transforms to guide mesh editing. We remove/add mesh parts according to the difference between the original and transformed point clouds. The mesh parts are retrieved from the ShapeNet part dataset~\cite{yi2016scalable} by Chamfer distance. Details of retrieval and more examples can be found in the supplementary material.

\begin{figure}[t!]
	\centering
	\includegraphics[width=\linewidth]{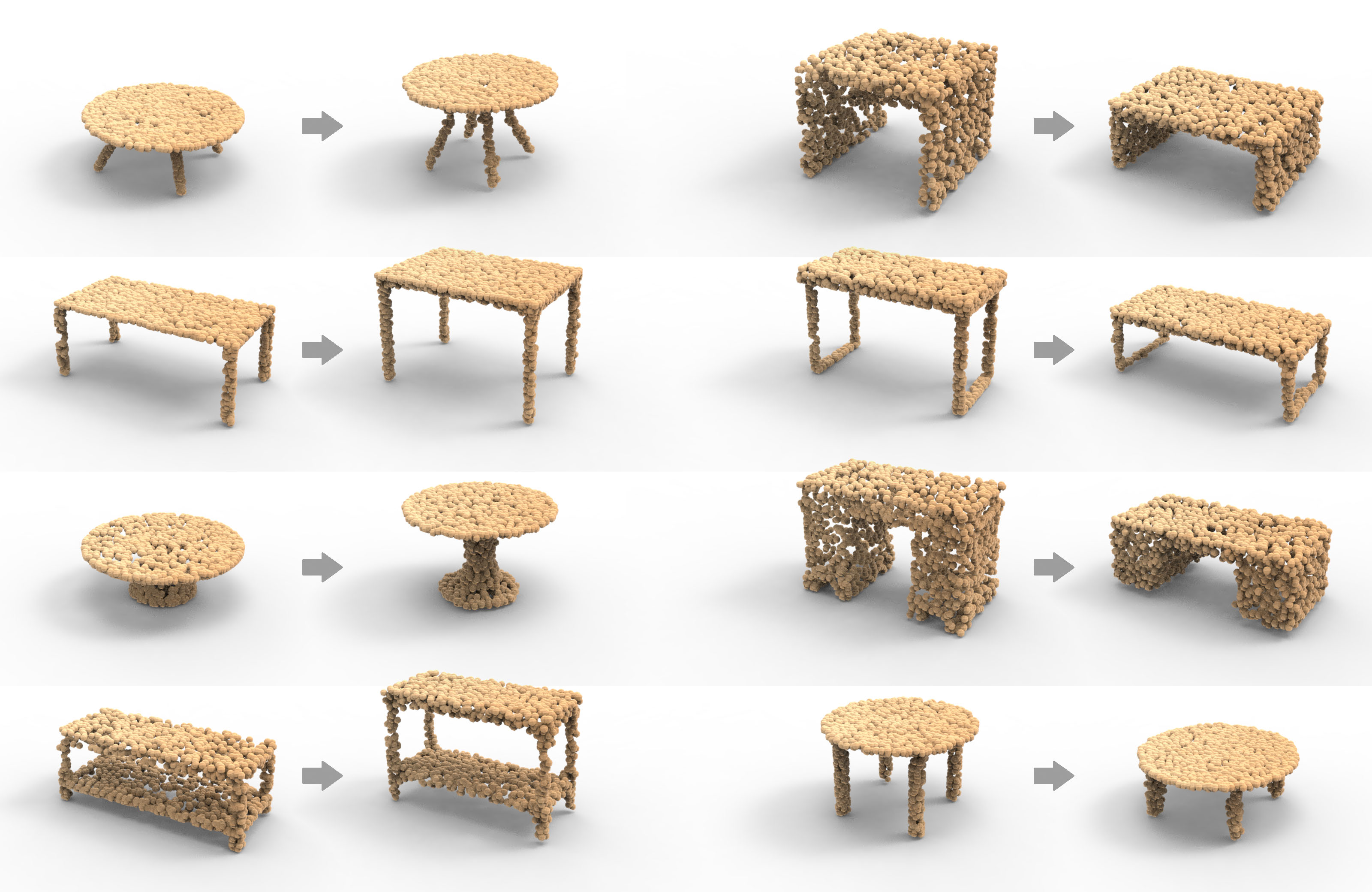}
	\caption{Unpaired shape transforms between tall and short tables. Left: increasing height. Right: decreasing height. 
        }
	\label{fig:tall_short}
\end{figure}

\paragraph{Transforming global 3D shape attribute.}
At last, we show that LOGAN is able to learn to increase/decrease heights of tables, which can be considered as a style for 3D shapes, as shown in Figure~\ref{fig:tall_short}. To obtain such a suitable dataset, we sort the height-width ratios of all the tables from ShapeNet Core dataset. The first 3,000 tables are selected as tall tables, while the last 3,000 as short tables. From each of the two sets, we randomly choose 500 shapes as the test set, and the rest are used for training. The results demonstrate that LOGAN is able to alter global attributes of 3D shapes. 

\subsection{Unpaired style/content transfer and comparisons}
\label{subsec:style_content}

\begin{figure}[t!]
	\centering
	\includegraphics[width=0.95\linewidth]{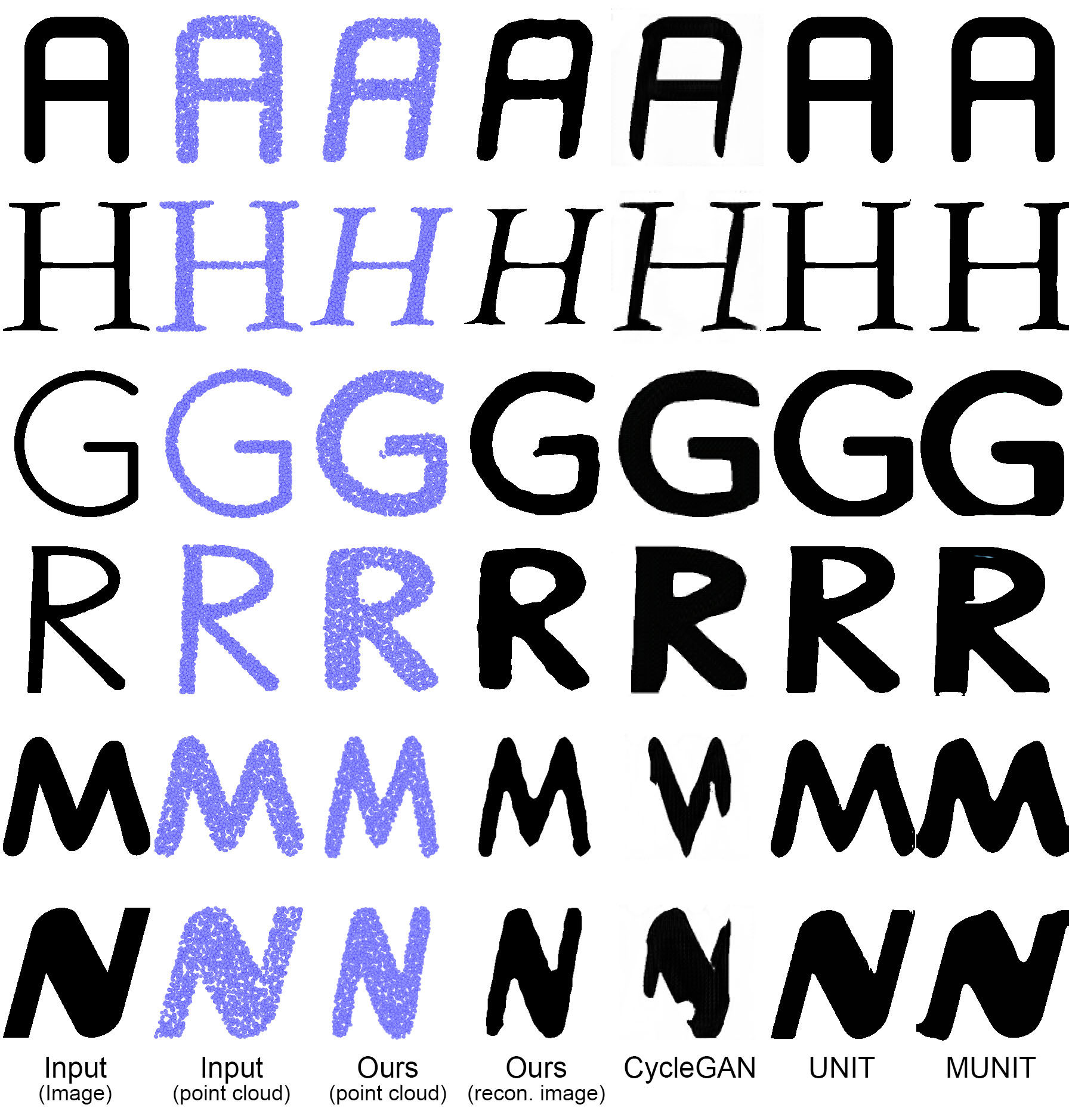}
	\caption{Comparisons on content-preserving style transfer, i.e., \textit{regularA/H-italicA/H}, \textit{thinG/R-thickG/R}, and \textit{wideM/N-narrowM/N} translations, by different methods. First two rows: regular-to-italic; middle two rows: thin-to-thick; last two rows: wide-to-narrow. From left to right: input letter images; corresponding input point clouds; output point clouds from LOGAN; images reconstructed from our results; output images of CycleGAN; outputs from UNIT~\cite{UNIT17}; outputs from MUNIT~\cite{MUNIT18}. For \textit{wideM/N-narrowM/N} we align the letters by height for better visualization.}
	\label{fig:fontStyleTranfer}
\end{figure}

\begin{figure}[t!]
	\centering
	\includegraphics[width=0.99\linewidth]{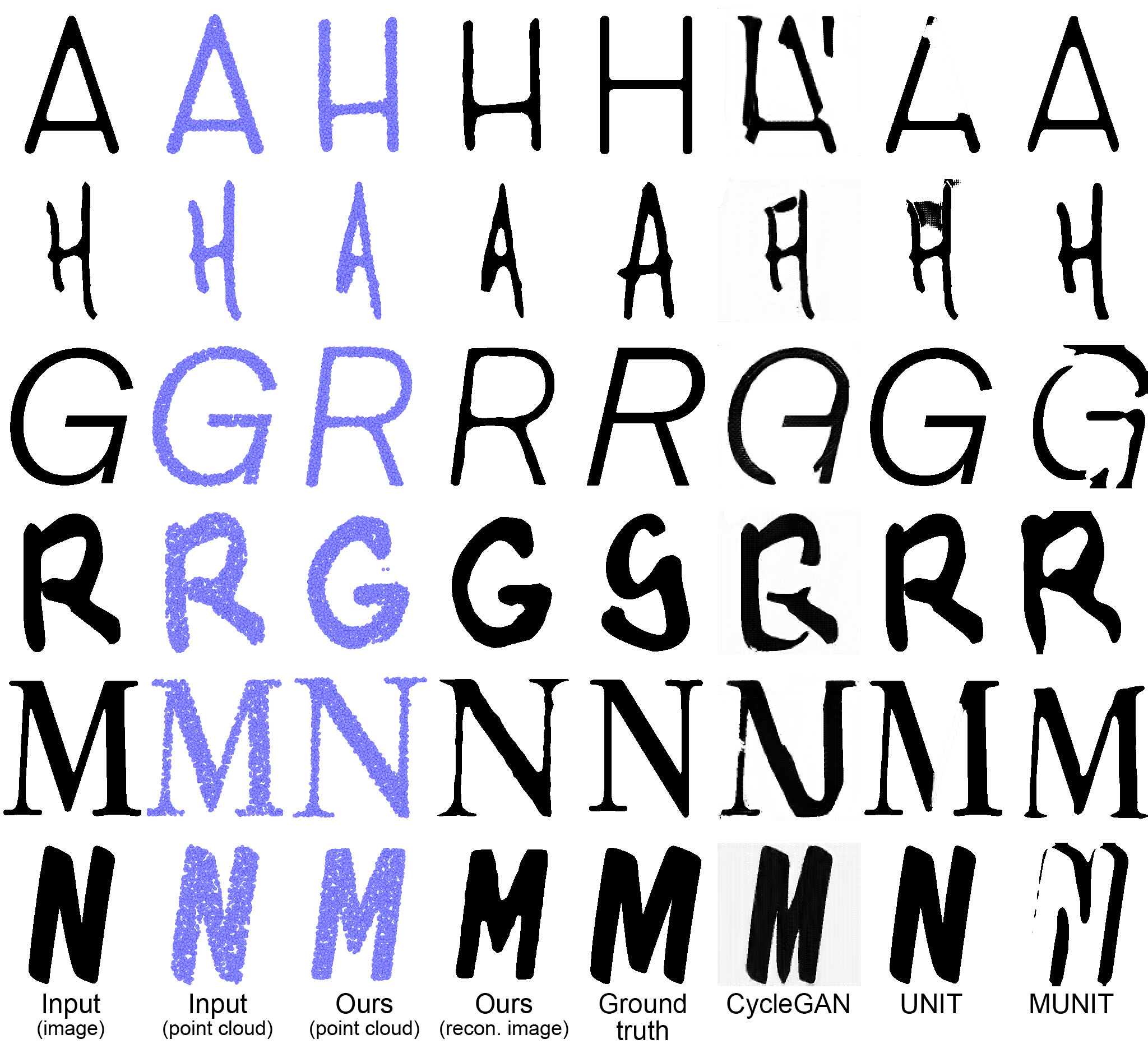}
	\caption{\revise{Comparisons on style-preserving content transfer, i.e., \textit{A-H}, \textit{G-R}, and \textit{M-N} translations, by different methods, including ground truth.}}
	\label{fig:G2R_compare}
\end{figure}

\begin{table}[t!]
\begin{center}
\begin{tabular}{l|r|r|r|r|r|r}
\hline
  & \multicolumn{2}{c|}{A$\leftrightarrow$H} & \multicolumn{2}{c|}{G$\leftrightarrow$R} & \multicolumn{2}{c}{M$\leftrightarrow$N} \\
\hline
 & MSE & IOU & MSE & IOU & MSE & IOU \\
\hline\hline
CycleGAN & 0.246 & 0.385 & 0.229 & 0.412 & 0.266 & 0.383 \\
UNIT & 0.253 & 0.376 & 0.264 & 0.377 & 0.295 & 0.348 \\
MUNIT & 0.280 & 0.286 & 0.358 & 0.171 & 0.363 & 0.292 \\
Ours & {\bf 0.195} & {\bf 0.490} & {\bf 0.213} & {\bf 0.472} & {\bf 0.207} & {\bf 0.506} \\
\hline
\end{tabular}
\end{center}
\caption{Quantitative comparisons on \textit{A-H}, \textit{G-R}, and \textit{M-N} translations by different unpaired cross-domain translation networks. Mean squared error (MSE) and intersection over union (IOU) are measured against ground-truth target letters and averaged over the testing split of the respective datasets. Better-performing numbers are highlighted in boldface.}
\label{table:G2R_numbers}
\end{table}

\begin{figure*}[!t]
	\centering
	\includegraphics[width=0.97\linewidth]{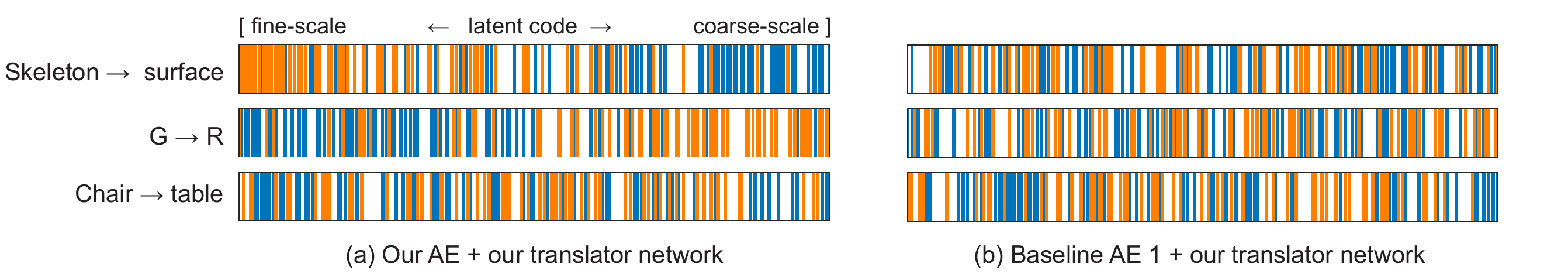}
	\caption{Visualizing ``disentanglement'' in latent code preservation (blue color) and alteration (orange color) during translation, where each bar represents one dimension of the latent code. Orange bars: top 64 latent code dimensions with the largest average changes. Blue bars: dimensions with smallest code changes.}
	\label{fig:code_viz}
\end{figure*}

Most deep networks for unpaired cross-domain translation operate on images, aiming for content-preserving style transfer~\cite{zhu2017unpaired,yi2017dualgan,UNIT17,MUNIT18}. We conduct an experiment to compare LOGAN with these state-of-the-art networks on a \textit{Font} dataset we collected. The dataset consists of 7,466 fonts of English letters. For each letter, we produce a rendered $256^2$ image by normalizing the letter to make the longest edge of its bounding box equal to $248$. Then we obtain a point cloud by uniformly sampling 2,048 points over the pixels inside each letter shape.



In the first test, {\em style transfer\/}, we train the networks to translate between different styles of various fonts, while keeping the source and target letters the same. In Figures~\ref{fig:fontStyleTranfer}, we show examples including \textit{regularA/H-italicA/H}, \textit{thinG/R-thickG/R}, and \textit{wideM/N-narrowM/N}.
Since most fonts in our collected dataset do not have paired regular and italic types, nor paired regular and boldface types, we split the dataset ourselves. To split the dataset for \textit{regularA/H-italicA/H}, we simply sort the fonts by checking how vertical the letter ``I'' is for each font. The first 2,500 fonts with more vertical ``I''s are regarded as \textit{regularA/H}s, while the last 2,500 as \textit{italicA/H}s. Similarly, we split the set of all $G$s and $R$s in the dataset simply by sorting them based on how many pixels are inside the letter shapes. The first 2,500 letters with more interior pixels are regarded as \textit{thickG/R}s, and the last 2,500 as \textit{thinG/R}s. To obtain the dataset for \textit{wideM/N-narrowM/N},  we sort $M$s and $N$s by looking at their width-height ratios. Finally, we randomly selected 500 fonts from each of the above sets to serve as test set for the specific tasks, while using the rest for training. An ideal translator would only change the specific style of an input letter while keeping the letter in the same font. 

The second comparison is on style-preserving {\em content transfer\/}, via letter translation tasks on three subsets: \textit{A-H}, \textit{G-R}, and \textit{M-N}. To obtain the \textit{G-R} dataset, from the 7,466 pairs of uppercase $G$s and $R$s, each in the same font, we randomly selected 1,000 to serve as the testing set while using the rest for training. The  datasets for \textit{A-H} and \textit{M-N} are obtained in the same way. For these task, we expect an ideal translator to transform samples between the two domains by changing the letter (content) only, while preserving its style, e.g., font width, height, thickness, skewness, etc.

We compare LOGAN with three unpaired image translation networks: the original CycleGAN~\cite{zhu2017unpaired} which translates images directly, as well as UNIT~\cite{UNIT17} and MUNIT~\cite{MUNIT18}, both of which utilize shared latent spaces. While LOGAN operates on point clouds, the other networks all input and output images. We trained each of the four networks for 15 hours for each of the \textit{regularA/H-italicA/H}, \textit{thinG/R-thickG/R}, and \textit{wideM/N-narrowM/N} translations; and 27 hours for each of the \textit{A-H}, \textit{G-R}, and \textit{M-N} translations. To help with comparison, we convert the output point clouds from LOGAN to images: for each point in a given point cloud, we find all its neighbors within $r$ pixels away, and then fill the convex hull of these points; we used $r=10$ in our tests.

Results for content-preserving style translation are shown in Figures~\ref{fig:fontStyleTranfer}, while Figure~\ref{fig:G2R_compare} compares results for style-preserving content transfer. More results can be found in the supplementary material.
We observe that CycleGAN, UNIT, and MUNIT are unable to learn transforms between global shape structures, which are necessary for letter translations and certain style translations.  
 
Overall, our network can adaptively learn which features (content vs.~style) to preserve and which to transfer according to the training domain pairs. We also provide quantitative comparisons in Table~\ref{table:G2R_numbers} for letter translations since we have ground-truth target letters. These results again demonstrate the superiority of LOGAN.


\subsection{Implicit disentanglement over latent codes}
\label{subsec:imp_dis}

We examine how our latent space representations may help disentangle preserved vs.~altered shape features during cross-domain translation. In Figure~\ref{fig:code_viz}, we plot latent code dimensions with the largest (top 64 out of 256 dimensions, in orange color) and smallest changes (bottom 64, in blue color) for three translation tasks: airplane skeleton$\rightarrow$ surface (Section~\ref{subsec:comp_p2p}), \textit{G}$\rightarrow$\textit{R}, and chair$\rightarrow$table. \revise{Note that the plots reflect the {\em mean\/} magnitude of 
code changes in each dimension, over the respective {\em whole test sets\/}.} 

We can observe that our network automatically learns the right features to preserve (e.g., more global or coarser-scale features for airplane skeleton$\rightarrow$ surface and more local features for \textit{G}$\rightarrow$\textit{R}), solely based on the input domain pairs. For the chair$\rightarrow$table translation, coarse-level and fine-level features are both impacted. Compared to PointNET++ encoding \rz{(baseline AE 1)}, our default autoencoder with overcomplete codes better disentangles parts of the latent codes that are preserved vs.~altered\rz{--- this is more pronounced for the first two examples and less so for the chair-table translation.}

\begin{table*}[t]
	\begin{center}
		\begin{tabular}{l|r|r|r|r|r|r}
			\hline
			& \multicolumn{2}{c|}{Skeleton $\leftrightarrow$ Shape} & \multicolumn{2}{c|}{Scan $\leftrightarrow$ Shape} & \multicolumn{2}{c}{Profiles $\leftrightarrow$ Surface} \\ \hline
			& Chamfer & EMD$/n$ & Chamfer & EMD$/n$ & Chamfer & EMD$/n$ \\ \hline
			PointNET++ autoencoder (AE) + Our translator with all three losses & 5.30 & 0.071 & 5.30 & 0.079 & 9.29 & 0.099 \\
			Our AE ($\lambda_1=0$) + Our translator with all three losses  & 2.24 & 0.048 & 2.42 & 0.051 & {\bf 3.08} & {\bf 0.061} \\
			Our AE ($\lambda_1=0.1$) + Our translator with only WGAN + Cycle losses & 14.06 & 0.098 & 16.74 & 0.127 & 17.06 & 0.116 \\
			Our AE ($\lambda_1=0.1$) + Our translator with only WGAN + FP losses & 2.22 & 0.047 & 2.53 & 0.054 & 3.37 & 0.064 \\
			{\bf LOGAN:\/} Our AE ($\lambda_1=0.1$) + Our translator with all three losses & {\bf 2.11} & {\bf 0.046} & {\bf 2.18} & {\bf 0.048} & 3.09 & {\bf 0.061} \\ \hline
			P2P-NET: Supervised method with paired domains & 0.44 & 0.020 & 0.66 & 0.060 & 1.36 & 0.056 \\ \hline
		\end{tabular}
	\end{center}
	\caption{Quantitative comparisons between different autoencoder and translator configurations, on transformation tasks from P2P-NET. Reported errors are averaged over two categories per domain pairs (see Figure~\ref{fig:comp_p2p}), and are measured against ground-truth target shapes from the P2P-NET dataset. Since P2P-NET did not come with an upsampling layer, for a fair comparison, all point cloud results are obtained at the same resolution of 2,048 points.}
	\label{table:quant_p2p}
\end{table*}

\begin{figure*}[t!]
	\centering
	\includegraphics[width=0.99\linewidth]{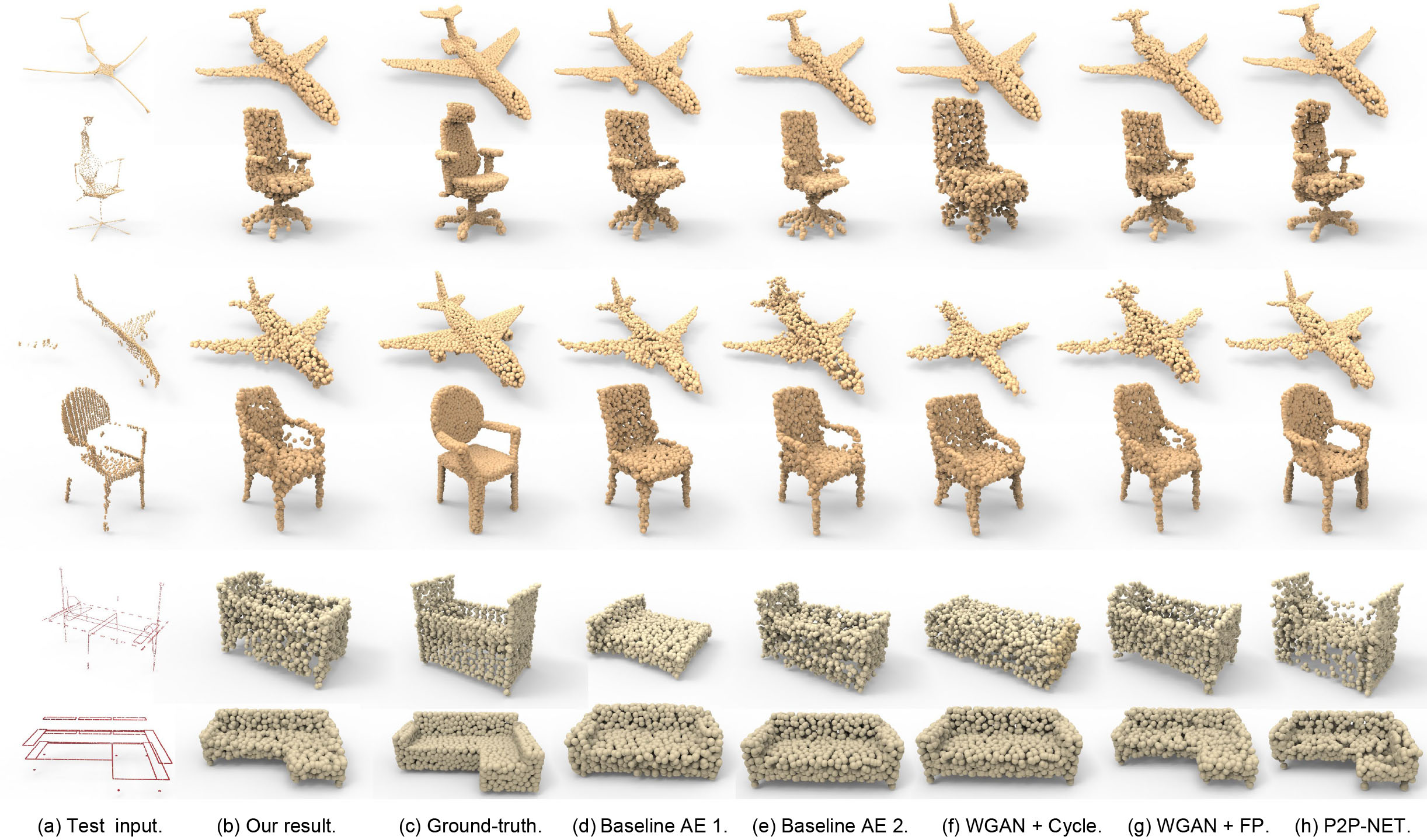}
	\caption{Comparisons between various network configurations, (supervised) P2P-NET, and ground truth targets, on shape transform examples from P2P-NET: skeleton$\rightarrow$shape (rows 1-2), scan$\rightarrow$surface (rows 3-4), and (cross-sectional) profiles$\rightarrow$surface (rows 5-6). All point clouds have 2,048 points.}
	\label{fig:comp_p2p}
\end{figure*}

\subsection{Comparison with supervised P2P-NET}
\label{subsec:comp_p2p}

In Figure~\ref{fig:comp_p2p}, we show unpaired cross-domain shape transform results obtained by LOGAN, on several domain pairs from the recent work P2P-NET~\cite{yin2018p2p}, where the specific input shapes are also from their work. We compare these results to P2P-NET, as well as results from other network configurations as done in Figure~\ref{fig:comp_chairtable}. Note that these shape transforms, e.g., cross-sectional profiles to shape surfaces, are of a completely different nature compared to table-chair translations. Yet, our network is able to produce satisfactory results, as shown in column (b), which are visually comparable to results obtained by P2P-NET, a supervised method. 

Both LOGAN and P2P-NET aim to learn general-purpose cross-domain transforms between point-set shapes. P2P-NET works on paired domains but without explicit feature preservation, while LOGAN is trained on unpaired data but enforces a feature preservation loss in the GAN translator. The results show that LOGAN is able to preserve the right global features for skeleton/scan-to-shape translations. At the same time, some finer details, e.g., the swivel chair legs and the small bump near the back of the fuselage in row 1, can also be recovered. However, the unsupervised LOGAN cannot quite match P2P-NET in this regard; see the back of the swivel chair.

Since P2P-NET is supervised, ground-truth target shapes are available to allow us to quantitatively measure the approximation quality of the translation results. As shown in Table~\ref{table:quant_p2p}, our default LOGAN network achieves the best quality, compared to other baseline alternatives, but still falls short of the supervised P2P-NET.




\subsection{Comparison with unpaired deformation transfer}
\label{subsec:comp_deftrans}

\begin{figure}[t!]
	\centering
	\includegraphics[width=0.99\linewidth]{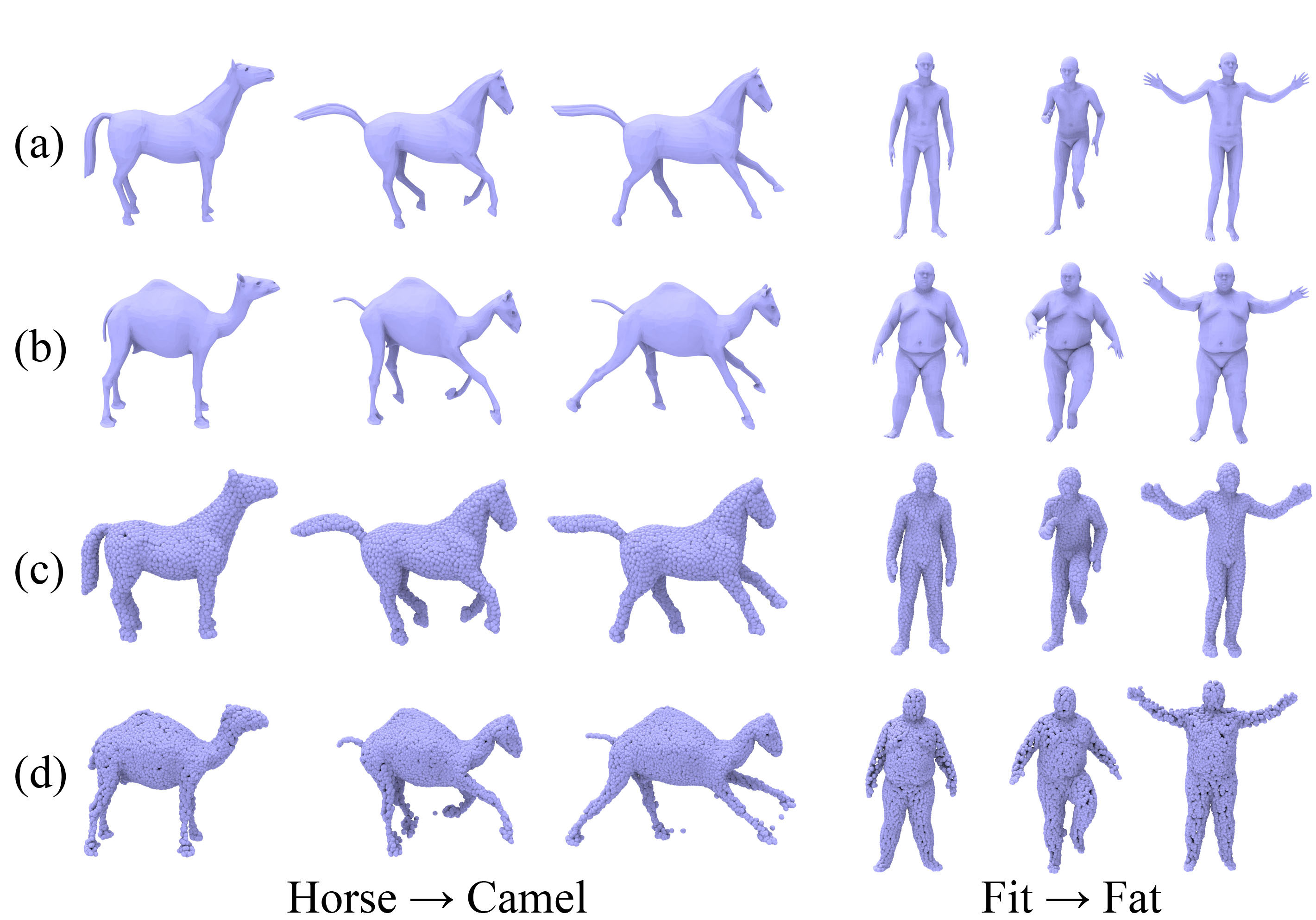}
	\caption{Comparison with unpaired deformation transfer~\cite{gao2018automatic}, demonstrating that LOGAN can also accomplish the task. (a) Input meshes; (b) output meshes by Gao et al.~\shortcite{gao2018automatic}; (c) input point clouds of size 2,048 sampled from (a); (d) output point clouds upsampled to 8,192 points.}
	\label{fig:shape_deformation}
\end{figure}

The latent VAE-CycleGAN developed by Gao et al.~\shortcite{gao2018automatic} was designed for the specific task of unpaired deformation transfer on meshes. In this last experiment, we further test the generality of our shape transformation network, by training it on datasets from~\cite{gao2018automatic}. In Figure~\ref{fig:shape_deformation}, we compare results obtained by LOGAN and results from~\cite{gao2018automatic} as provided by the authors. The point clouds for training were obtained by uniformly sampling 2,048 points from the corresponding meshes.  Note that
the \textit{horse $\to$ camel} dataset contains a total of 384 shapes in the training set; \textit{fit $\to$ fat} contains 583 shapes. Since these datasets are significantly smaller than those from the previous experiments, we adjusted the hyperparameter $\lambda_2$ to $40$ in order to better avoid overfitting, and increase the number of training epochs for translators to 1,200.

The results show that, {\em qualitatively\/}, LOGAN, which is designed as a general-purpose cross-domain shape transform network, is also able to learn to preserve pose-related features and achieve pose-preserving shape transform, like Gao et al.~\shortcite{gao2018automatic}. However, since our current implementation of the network is limited in training resolution (at 2,048 points), the visual quality of the generated point clouds does not quite match that of their mesh outputs.

%
%

\section{Discussion, limitation, and future work}
\label{sec:future}

\begin{figure}[t!]
	\centering
	\includegraphics[width=0.9\linewidth]{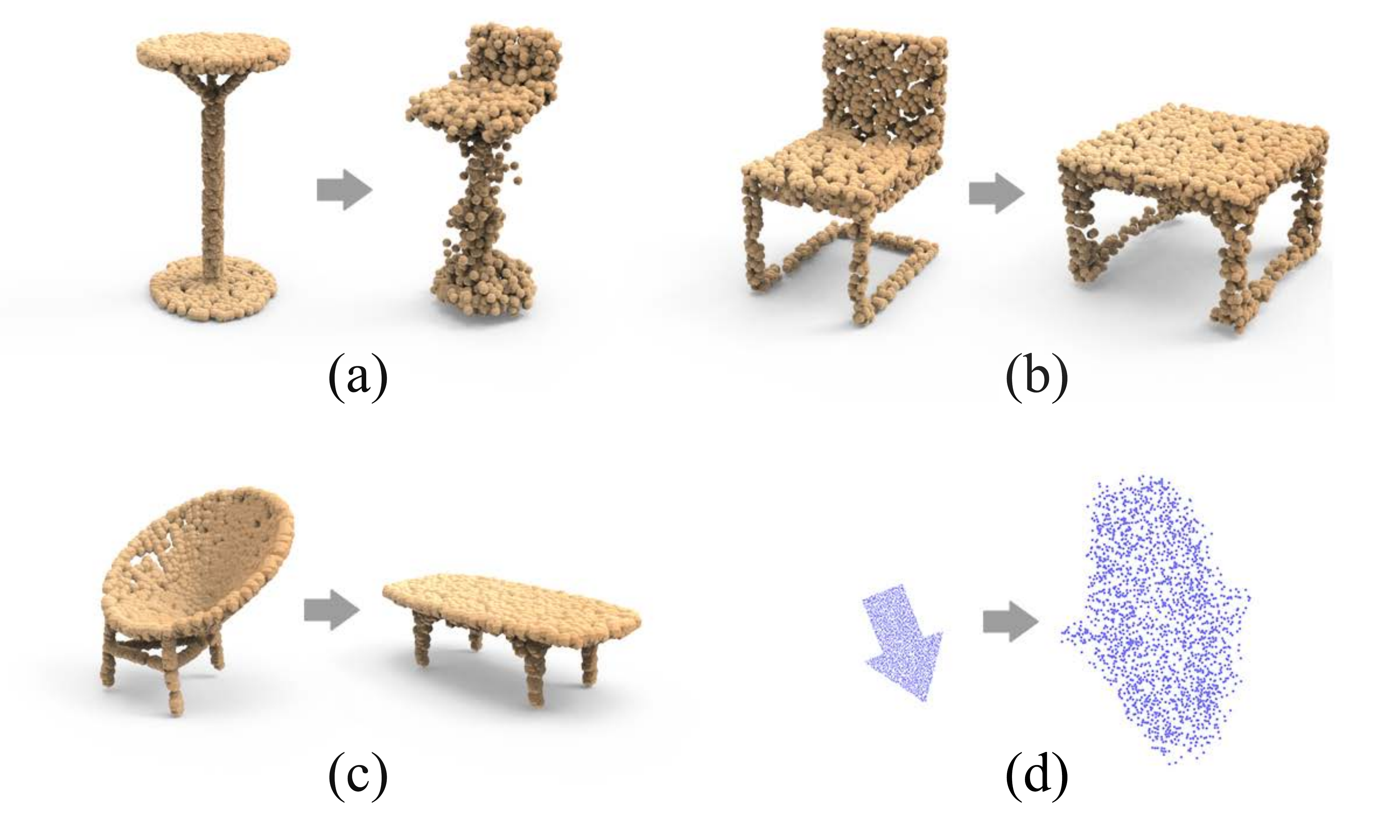}
	\caption{Several failure cases by LOGAN: (a) scattered point clouds; (b-c) unnatural transform results due to lack of commonalities in the target domain; (d) failed translation between shapes differing in global scales.}
	\label{fig:failcase}
\end{figure}

Shape transform, or shape-to-shape translation, is a basic problem in geometric modeling. We believe that it is as fundamental to computer graphics as image-to-image translation is to computer vision.
We develop a deep neural network for learning generic cross-domain shape translations. The key challenge posed is how to ensure that the network can learn shape transforms between two domains {\em without\/} any paired shapes. Our motivation is that for most modeling tasks, especially those involving 3D models, it is difficult to find pre-existing paired data due to a lack of 3D models to begin with. On the other hand, synthesizing paired 3D data is unrealistic since 3D modeling is generally a non-trivial task; this is the very reason why we seek learning methods to automate the process. 


Our ultimate goal is to have machine learning assist in any {\em transformative\/} design tasks carried out by a graphics artist/modeler, where there are no ground truth targets, only a target domain as inspiration. LOGAN only makes a first step towards this direction, laying a foundation for follow-up research.

Cross-domain shape translation is not suitable for all domain pairs, e.g., (chair, airplane) or (human body, face). Hence, there is an implicit assumption that shapes from the two domains should share some commonalities. In general, these commonalities may be latent; they may reflect global or local features and represent either content or style. The key is for the network to learn the right commonalities and keep them during shape translations, in a way that is adaptive to the input domain pairs and the input shapes. Our network is designed with several important features to accomplish this: autoencoding shapes from two domains into a common latent space; the feature preservation loss; and perhaps most importantly, the use of multi-scale and overcomplete latent codes. 


Another aspect of our work is the separation of autoencoder training from the latent cross-domain translation network. Unlike most previous works on unpaired cross-domain image translation, e.g.,~\cite{UNIT17,MUNIT18}, we do not use a combined loss. Similar to Gao et al.~\shortcite{gao2018automatic}, we train the autoencoder and translator separately. We also believe that the separation facilitates training of the GAN translators and leads to improved results.  

We regard our work as only making a first step towards generic, unpaired cross-domain shape transform, and it still has quite a number of limitations. First, due to the inherent nature of point cloud representations, the output shapes from our network are not necessarily clean and compact. The points can be scattered around the desired locations, especially when there are thin parts; see Figure~\ref{fig:failcase}(a) and some results in Figure~\ref{fig:G2R_compare}. 

Second, due to our assumption of shared commonalities between the input domains, if an input shape in one domain cannot find sufficient commonalities in the other domain, our network cannot learn a natural translation for it; see Figures~\ref{fig:failcase}(b-c). In such cases, the adversarial loss plays a more dominant role. We can observe the impact of this loss in rows 1-3 of Figure~\ref{fig:comp_chairtable}. These chair$\rightarrow$table translation results by LOGAN cannot retain the squared tops, which may be judged by some as an unnatural transform; the reason is that most tables in the training set have rectangular tops. Similarly, the result in Figure~\ref{fig:failcase}(b) is not a complete failure as the output table did preserve the square top as well as certain leg features. Simply removing the chair back would result in an unusual table.

Third, performing translations in a common space and measuring the feature preservation loss entry-by-entry imply that we implicitly assume a ``scale-wise alignment'' between the input shapes. That is, the common features to be preserved should be in the same scales. Figure~\ref{fig:failcase} (d) shows a result from LOGAN which was trained to translate between arrow shapes of very different scales; the result is unnatural due to a lack of that scale-wise alignment.
\rz{Last but not least, as a consequence of employing latent space transforms, our method is unable to output shape correspondences between the source and target shapes as 
prior works could.} 


In future work, we would like to consider other overcomplete, concatenated shape encodings where the different representations reflect other, possibly semantic, aspects of the shapes, beyond their multi-scale features. We would also like to expand and broaden the scope of shape transforms to operations such as shape completion, style/content analogy, and more. Finally, semi-supervision or conditional translations~\cite{MUNIT18} to gain more control on the transform tasks are also worth investigating.

%
%


\section*{Acknowledgments}
The authors would like to thank the anonymous reviewers for their valuable comments. Thanks also go to Haipeng Li and Ali Mahdavi-Amiri for their discussions, and Akshay Gadi Patil for proofreading. This work was supported by NSERC Canada (611370), gift funds from Adobe, NSF China (61761146002,61861130365), GD Science and Technology Program (2015A030312015), LHTD (20170003), ISF (2366/16), and ISF-NSFC Joint Research Program (2472/17).

\bibliographystyle{ACM-Reference-Format}
\bibliography{points}

\end{document}